\documentclass[journal]{IEEEtran}

\usepackage{cite}
\usepackage[pdftex]{graphicx}
\usepackage{amsmath}
\usepackage{subfigure}
\usepackage{url}
\usepackage{booktabs,bigstrut,multirow,color}
\begin{document}
\title{MFFW: A new dataset for multi-focus image fusion}

\author{Shuang~Xu,
        Xiaoli~Wei,
        Chunxia~Zhang,
        Junmin~Liu,~\IEEEmembership{Member,~IEEE,}
        and~Jiangshe Zhang,% <-this % stops a space
\thanks{S. Xu, X. Wei, C. Zhang, J. Liu and J. Zhang are with School of Mathematics and Statistics, Xi'an Jiaotong University, Xi'an, Shaanxi Province, 710049 China e-mail: (shuangxu@stu.xjtu.edu.cn; wxl18847162006@stu.xjtu.edu.cn; cxzhang@mail.xjtu.edu.cn; jmliu@mail.xjtu.edu.cn; jszhang@mail.xjtu.edu.cn).}% <-this % stops a space
\thanks{Manuscript received Jan. 29, 2020; revised XX XX, 2020.}}

\markboth{Journal of \LaTeX\ Class Files,~Vol.~14, No.~8, August~2020}%
{Shell \MakeLowercase{\textit{et al.}}: Bare Demo of IEEEtran.cls for IEEE Journals}



\maketitle

\begin{abstract}
Multi-focus image fusion (MFF) is a fundamental task in the field of computational photography. Current methods have achieved significant performance improvement. It is found that current methods are evaluated on simulated image sets or Lytro dataset. Recently, a growing number of researchers pay attention to defocus spread effect, a phenomenon of real-world multi-focus images. Nonetheless, defocus spread effect is not obvious in simulated or Lytro datasets, where popular methods perform very similar. To compare their performance on images with defocus spread effect, this paper constructs a new dataset called MFF in the wild (MFFW). It contains 19 pairs of multi-focus images collected on the Internet. We register all pairs of source images, and provide focus maps and reference images for part of pairs. Compared with Lytro dataset, images in MFFW significantly suffer from defocus spread effect. In addition, the scenes of MFFW are more complex. The experiments demonstrate that most state-of-the-art methods on MFFW dataset cannot robustly generate satisfactory fusion images. MFFW can be a new baseline dataset to test whether an MMF algorithm is able to deal with defocus spread effect. 
\end{abstract}


\IEEEpeerreviewmaketitle

\section{Introduction}
\IEEEPARstart{D}{ue} to the limitation of imaging devices, it is unable to keep all planes of a scene in focus. Consequently, only the objects in focus are sharp, while those not in focus are blurred. Multi-focus image fusion (MFF) is one of the promising techniques to obtain all-in-focus images \cite{survey}. The working mechanism of MFF is to capture images with different focuses, and to generate a new image by blending all clear regions of source images. Generally speaking, there are two categories of MFF technique, that is, classic and deep learning (DL) based methods. 

The classic methods can be further divided into two groups. The first one is the transform based method \cite{DSIFT,MSTSR,wavelet,nsct2,CSR,SR}. Its key idea is to convert the image from image space into feature space so as to make the active features are easily detected. And then, the all-in-focus image is reconstructed from feature space into image space after merging the active features according to a fusion strategy. The popular transformers include discrete wavelet transform \cite{wavelet}, non-subsampled contourlet transform \cite{nsct,nsct2} and sparse representation \cite{CSR,SR,7398058}. However, this kind of methods are often criticized, because they convert images into the handcraft feature spaces by incorporating certain prior knowledge. It remains a question that whether they can deal with various cases. The second group is the spatial domain based method \cite{BF,BRWTSFM,CBF,GFDF,GFF,MWGF}. Different from the first group, the second one directly operates on original images. This group can be further categorized into two classes, namely, filtering and segmentation. Filtering based methods \cite{CBF,GFDF,GFF} compute the active levels in local blocks, and apply the weighted average strategy to image fusion. Segmentation based methods \cite{BF,BRWTSFM,MWGF,8125753} segment source images based on the detected focus map, and combine in-focus objects from different source images. Nonetheless, the performance of this group strongly depends on the accuracy of detected focus map or active level.

\begin{figure}
	\centering
	\subfigure[]{\includegraphics[width=0.24\linewidth]{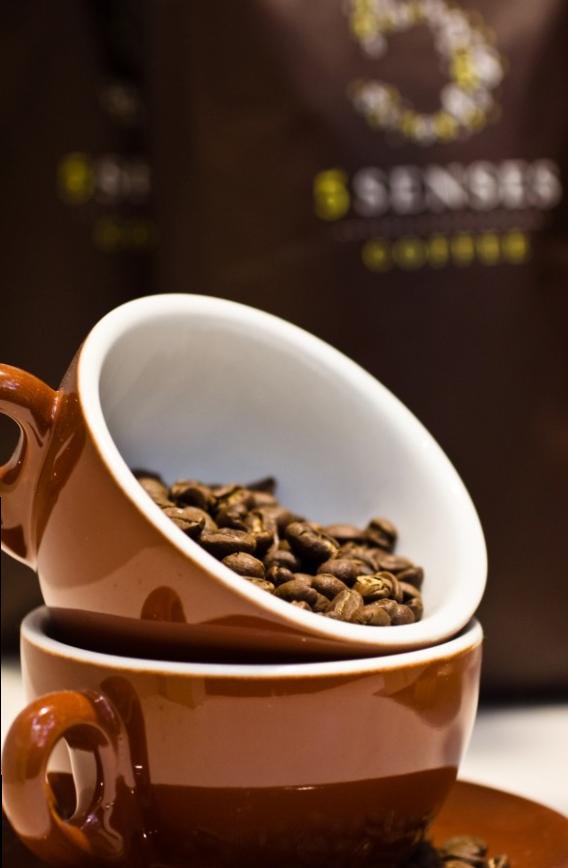}}
	\subfigure[]{\includegraphics[width=0.24\linewidth]{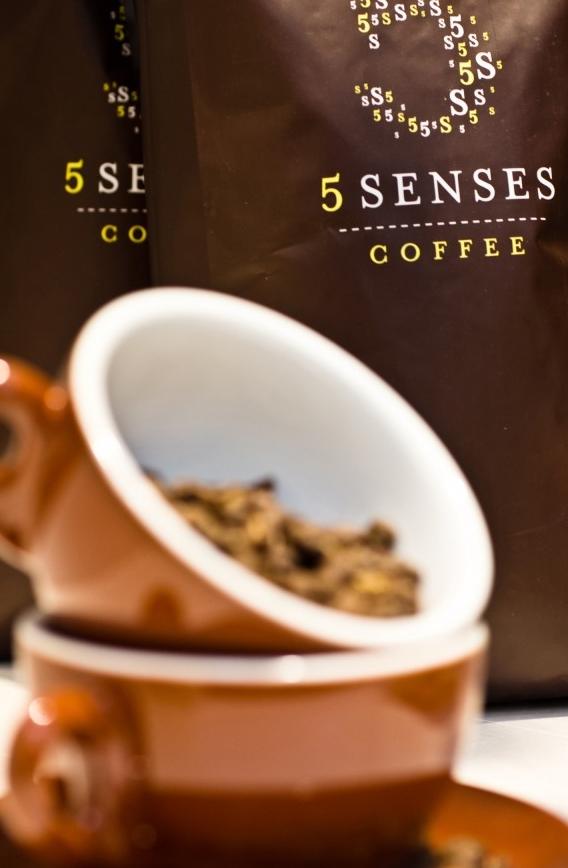}}
	\subfigure[]{\includegraphics[width=0.24\linewidth]{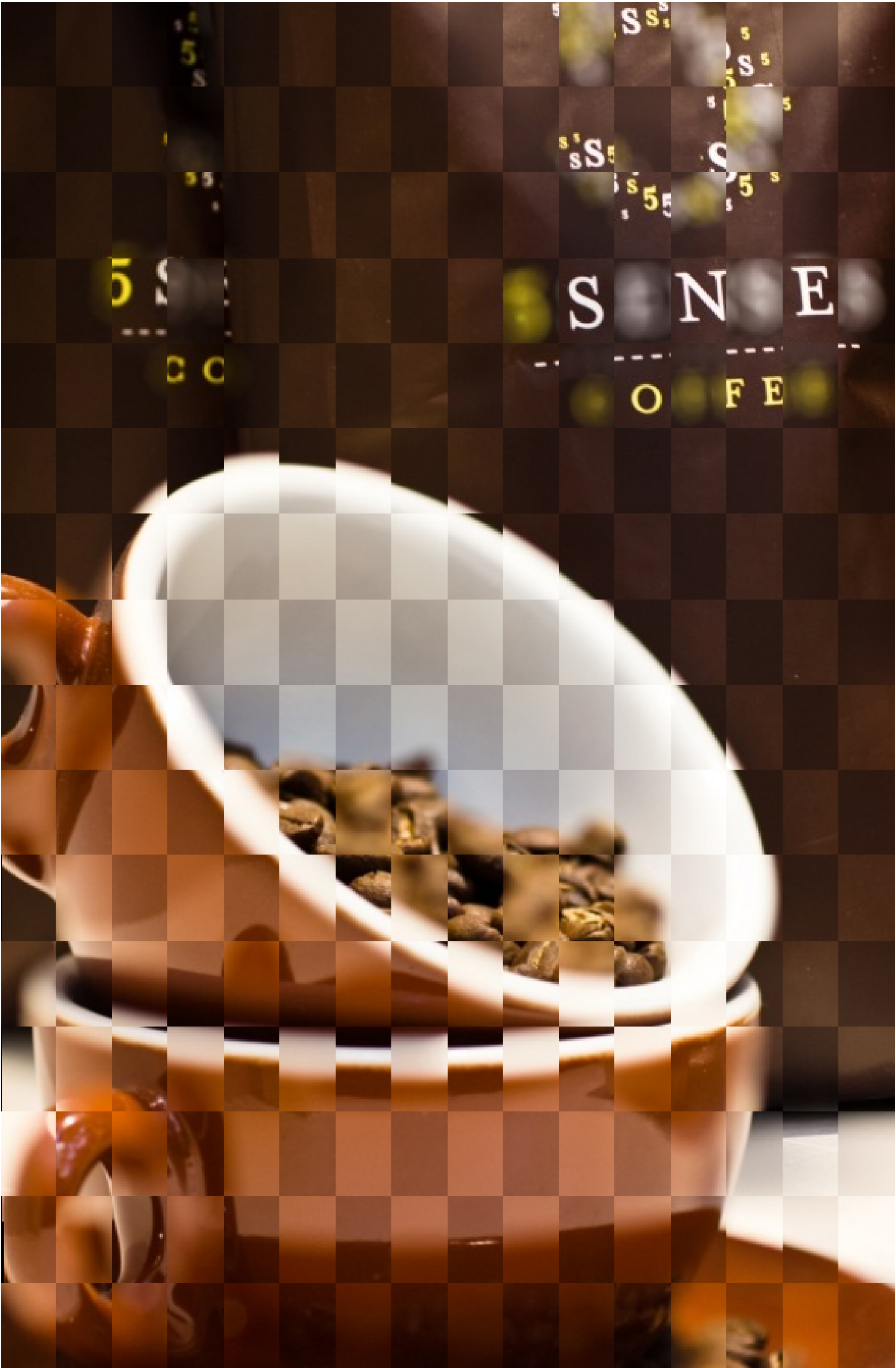}}
	\subfigure[]{\includegraphics[width=0.24\linewidth]{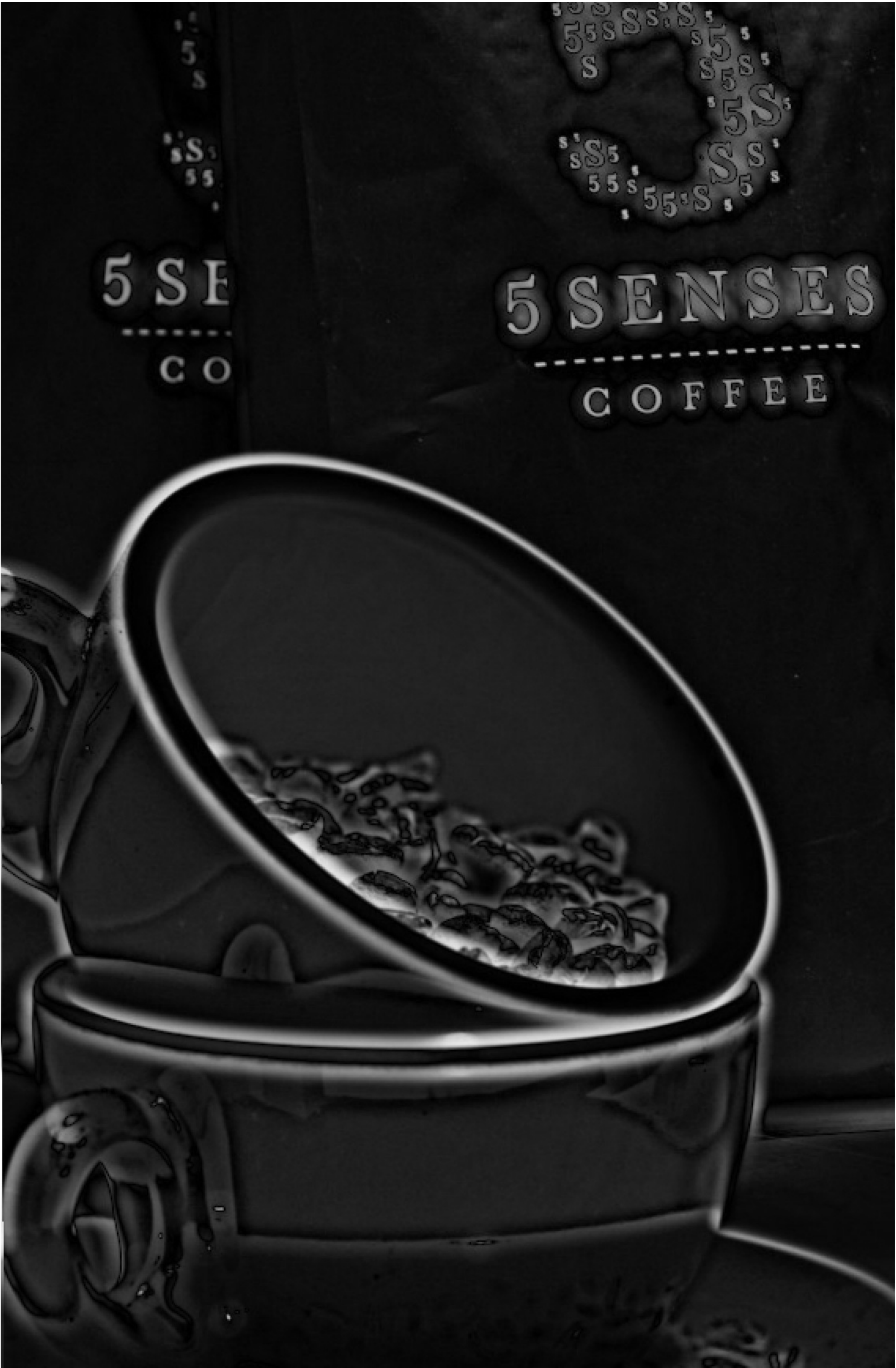}}
	
	\caption{Illustration for DSE. (a) Foreground focus. (b) Background focus. (c) Checkerboard with alternating rectangular regions from source images. (d) Difference image.}
	\label{fig:image1}
\end{figure}

The past decades have witnessed the fast development of DL \cite{DL}. An increasing number of researchers start to employ deep neural networks to fuse multi-focus images. The DL-based MFF can be categorized into two classes. In the first class, it trains an autoencoder which is the data-driven transformer. In the testing phase, it blends the feature maps (or say, codes) generated by the encoder, and fusion image is recovered by the decoder \cite{SESF,IFCNN}. Another class is based on supervised DL. In general, given an all-in-focus image and corresponding focus map, it simulates multi-focus images by a predefined defocus model. Since the ground-truth focus map is available, it is easy to train a pixel level classification network \cite{CNN,MFF_NET,PCNN,pena2019multiple,fusegan}. Obviously, this class can be also regarded as the focus map detection based method. By the virtue of DL, the focus map can be accurately detected.

Although DL has boosted the performance of MFF methods, it is reported that most of the algorithms suffer from the issue of unsatisfactory fusion results near focus boundary \cite{MFF_NET,Ma2019ICME}. As stated in recent references, regions far away from the focus boundary are totally focused or defocused, while there is the defocus spread effect (DSE) in areas near the focus boundary (see Fig. \ref{fig:image1}). According to this prior knowledge, references \cite{MFF_NET,Ma2019ICME} are devoted to training networks on the simulated datasets with DSE, intending to deal with focus boundary.

Note that most of the recent algorithms are tested on simulated image sets or a benchmark dataset, Lytro \cite{SR,Lytro}. However, it is worth to point out that both simulated and Lytro datasets to some degree cannot distinguish the better performer, because the DSE is not obvious on them. In much literature, it is found that there is no significant visual difference among fusion images generated by different algorithms. Only a few no-reference assessment metrics can show the minute disparity. But it has been pointed out that these metrics do not necessarily account for human visual system \cite{pena2019multiple}, and that in some cases these metrics may lead to a confused judgment \cite{Liu2012}. Therefore, it is urgent to propose a new dataset to evaluate whether these methods solve DSE or not.

\begin{figure}
	\centering
	{\includegraphics[width=1\linewidth]{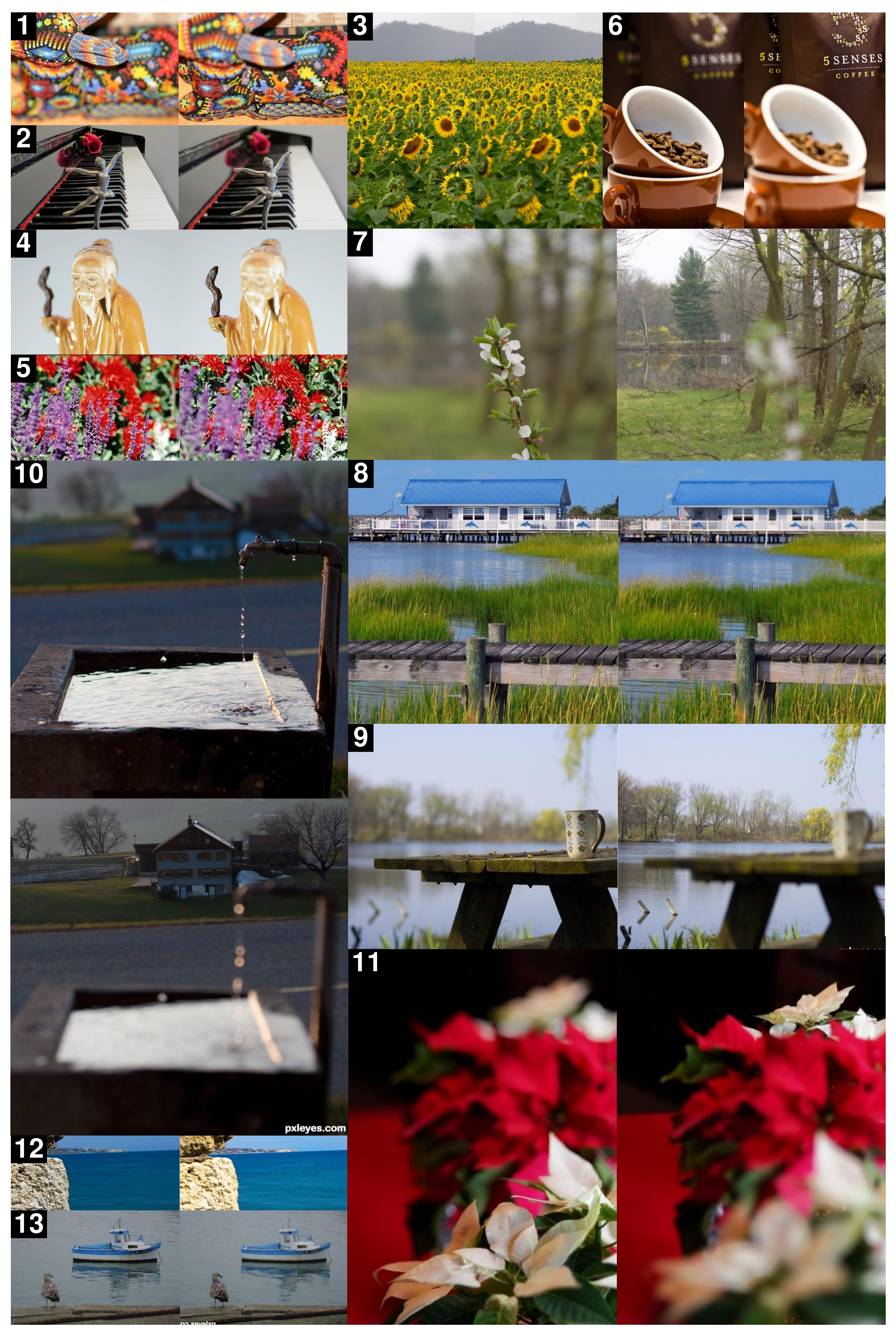}}
	\caption{Thirteen pairs of 2 source images in MFFW2.}
	\label{fig:image2}
\end{figure}

\begin{figure}
	\centering
	{\includegraphics[width=1\linewidth]{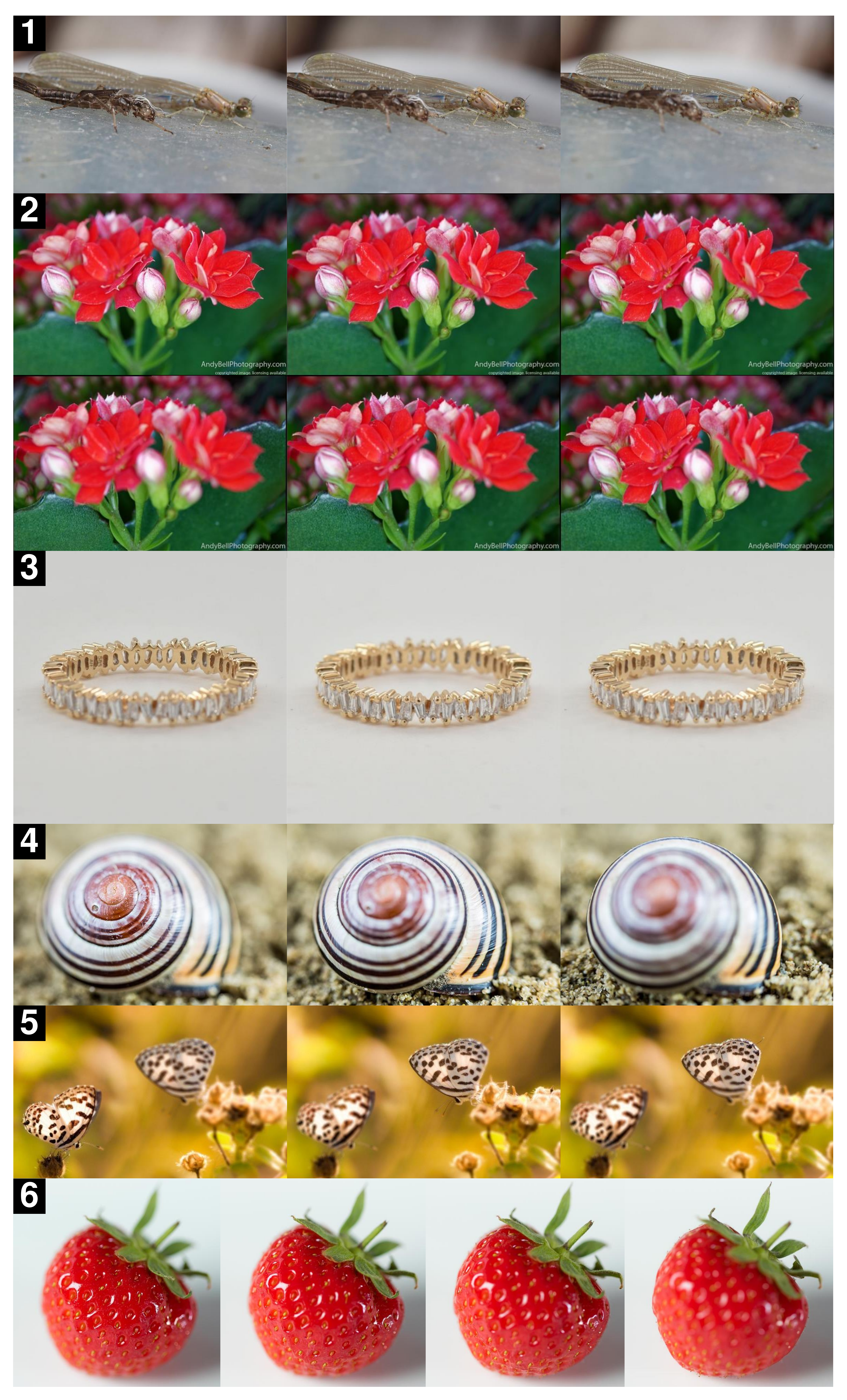}}
	\caption{Six pairs of 3+ source images in MFFW3.}
	\label{fig:image3}
\end{figure}

In this paper, to overcome the mentioned issue, we present a new testing dataset, called MFF in the wild (MFFW). MFFW consists of 13 pairs of two source images and 6 pairs of more than two source images, as shown in Figs. \ref{fig:image2} and \ref{fig:image3}. It is observed that images in MFFW are with evident DSE and complicated scenes. For two-source-images pairs, we provide annotated focus maps and manually edited reference images. We apply 13 state-of-the-art (SOTA) algorithms to MFFW dataset, including 9 classic methods and 4 DL based models. The visual inspection and 13 evaluation metrics demonstrate that none of algorithms on MFFW dataset can robustly generate satisfactory results. MFFW dataset can be a new baseline to evaluate whether an MMF algorithm is able to deal with DSE. 


\section{Main contribution}
In this section, we first state the motivation of our work, and then we introduce peculiarities of the novel dataset.

\subsection{Motivation}
The motivations of this work are threefold.

\subsubsection{Defocus spread effect}
As stated in Introduction section, one of obstacles to MFF methods is the DSE, that is, the objects which are far away from the focus tend to be larger than the original focused object. For ease of illustration, a pair of multi-focus images is displayed in Fig. \ref{fig:image1}. The DSE can be easily observed in Fig. \ref{fig:image1} (c) and (d). When the foreground is focused, characters in coffee bags significantly expand; in the contrast, cups expand when the background is focused. 

DSE leads to two issues. Firstly, empirical studies demonstrate that focus map detectors often make mistakes around the edges of expanded objects. Thereafter, the fusion image will be unrealistic and unnatural. Secondly, due to the depth of field, there is a blurred area in both source images. Namely, when background is focused, the background objects near the focus boundary are covered by blurred and expanded foreground objects. It is difficult for MFF methods to inpaint this area.

Nonetheless, it is found that the DSE is not very strong in Lytro dataset. Numerical experiments conducted on Lytro dataset show that recent models are able to deal with slight DSE, but can they keep the same performance when source images are corrupted by strong DSE? It remains a question.

\subsubsection{Imaging device}
The images in Lytro dataset \cite{SR,Lytro} are captured by Lytro camera \cite{georgiev2013lytro}. It is well-known that Lytro is a plenoptic camera which uses a microlens array as an imaging system focused on the focal plane of the main camera lens \cite{Lytro}. Besides Lytro cameras, there are different kinds of cameras in the consumer markets, such as digital single-lens reflex cameras and mirrorless interchangeable-lens cameras. So, it is very interesting to investigate the performance of MFF methods on images captured by other kinds of cameras.

\subsubsection{Shooting scenes}
Last but not least, it is found that the shooting scenes of images from Lytro dataset are relatively simple, so it is easy to detect focus maps. The experiments in later section will demonstrate that many focus map detection algorithm may lose efficiency if scenes become complicated.

\subsection{Peculiarities}
It is common knowledge that recent MFF methods have achieved satisfactory results on Lytro. In the last subsection, three limitations of Lytro dataset have been pointed out. Currently, it is interesting to investigate methods' performance on real-world multi-focus images captured by other imaging devices (rather than Lytro cameras) with relatively strong DSE or complex shooting scenes. 

To this end, we present a novel dataset called MFFW, consisting of 19 pairs of multi-focus images collected on the Internet. Note that 13 pairs contain two source images, while the other 6 pairs contain more than two source images. In what follows, we call 13 pairs of two source images as MFFW2 dataset, and 6 pairs of more than two source images as MFFW3 dataset. There are mismatches on all the pairs of source images. Hence, we register them by \textit{Fiji} \footnote{\url{https://imagej.net/Fiji/}} and \textit{Photoshop CS6} \footnote{\url{https://www.photoshop.com/}} softwares. It is worth to lay emphasis on several peculiarities of MFFW dataset. 

Firstly, all pairs are meticulously selected as shown in Figs. \ref{fig:image2} and \ref{fig:image3}. The DSE can be easily observed in many pairs including Nos. 1, 2, 4, 6, 7, 9, 10 and 11 in Fig. \ref{fig:image2}, and Nos. 2, 4, 5 and 6 in Fig. \ref{fig:image3}. Here, we take No. 10 in Fig. \ref{fig:image2} as an example. In this scene, a waterpipe and a pool are in near focus, while a cabin is in distant focus. It is evidently the water drops considerably expand when they are not in focus. The scenes of other pairs are relatively complex. For example, there is a pottery with gray background in No. 4 of Fig. \ref{fig:image2}. It is displayed that the old man and his walking stick are in near and distant focuses, respectively. The gray background is always out of focus. Intuitively, this scene is simple enough to identify the focus map accurately. Nevertheless, our experiment results are dramatically counter-intuitive. Actually, it is difficult to acquire satisfactory fusion images on our novel dataset. 

Secondly, ground-truth focus maps are provided for the pairs of two source images. We employ \textit{labelme} software \footnote{\url{https://pypi.org/project/labelme/}} to manually annotate focus maps with high-quality. As shown in Fig. \ref{fig:image2}, the sharper objects are carefully annotated by polygons. In our released dataset, the source annotation .json files are open access. The users can freely edit the annotated polygons if they want to. With the ground-truth focus maps, it is easy to assess the performance of focus map detectors.

Thirdly, as stated in last subsection, there are areas near focus boundary being blurred in both source images. This is a main crux for MFF task, since it is hard to recover the real scene with inadequate information. For ease of assessment, reference images are also provided in MFFW2 dataset, which are manually edited with \textit{Photoshop CS6} software (rather than the auto-blend layers option in \textit{Photoshop CS6}). Even though these reference images may be different from real scene, they conform with the human visual system. To certain degree, they can be regarded as auxiliary tools to assess the performance of MFF methods on vague areas.

\begin{figure}
	\centering
%	\subfigure[]{\includegraphics[width=0.49\linewidth]{image4a}}
%	\subfigure[]{\includegraphics[width=0.49\linewidth]{image4b}}
	\subfigure[]{\includegraphics[width=0.49\linewidth]{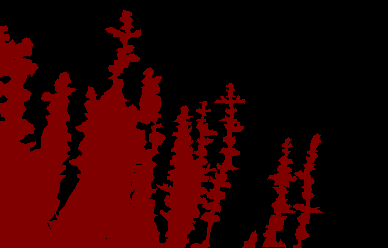}}
	\subfigure[]{\includegraphics[width=0.49\linewidth]{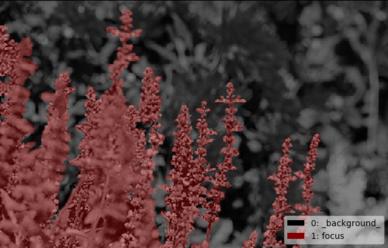}}
	\subfigure[]{\includegraphics[width=0.99\linewidth]{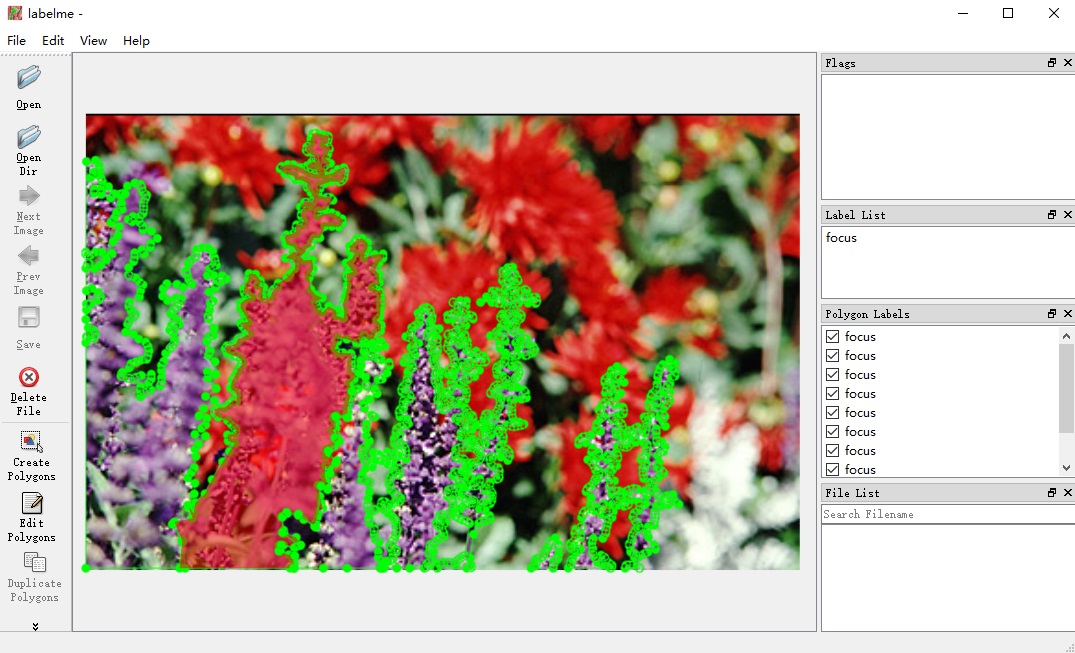}}
	\caption{An example of annotation. (a) Foreground focus. (b) Background focus. (c) Annotated focus map. (d) Blended image with focus map. (e) Interface of \textit{labelme}.}
	\label{fig:image4}
\end{figure}

\section{Numerical experiments}
This section will conduct experiments on MFFW dataset. The analysis on experimental results is also presented.

\subsection{Methods}
We test the performance of 9 classic methods and 4 deep learning based models on our dataset. As shown in Table \ref{tab:method}, they are representative and SOTA methods. All methods were implemented by official Matlab or Python codes and ran on a computer with Intel Core i7-9750H CPU@2.60GHz and RTX2080ti GPU. 
If a method is designed for gray-scale images, we separately apply it to R, G and B channels, and then combine them together as the final fused image. If a method is designed for two-source-images pairs, we apply it one by one in the context of more than two source images.

For DL based methods, the authors of CNN, IFCNN and SESF provide publicly available pre-trained networks which are used in their papers. However, since MMF-Net is trained on a publicly unavailable dataset, the authors of MMF-Net do not make the pre-trained network accessible. In this paper, we train MMF-Net on the NYU Depth V1 dataset\footnote{\url{https://cs.nyu.edu/~silberman/datasets/nyu_depth_v1.html}}. It is comprised of 2284 pairs of RGB and depth images. Note that the pixels of a depth image range from 0 to 1, representing the distance of objects from the viewpoint. To generate focus map, we portion the scene into focused and blurred regions according to a distance threshold, which is randomly sampled from interval $[0.2,0.6]$. With RGB images and corresponding focus maps, multi-focus images are simulated by the $\alpha$-matte boundary defocus model proposed by \cite{MFF_NET}. The training configuration of MMF-Net is the same as literature \cite{MFF_NET}. As for IFCNN, there are three different fusion strategies. Suggest by literature \cite{IFCNN}, we adopt max-fusion strategy in the context of MFF task. 
 
\begin{table}[tbp]
	\centering
	\caption{The information of MFF methods. SR and DL denote sparse representation and deep learning, respectively.}
\begin{tabular}{clll}
	\hline
	& Reference & Name  & Category \bigstrut\\
	\hline
	\multirow{9}[2]{*}{Classic methods} & \cite{BF} & BF    & Segmentation \bigstrut[t]\\
	& \cite{BRWTSFM} & BRWTSFM & Segmentation \\
	& \cite{CBF} & CBF   & Filtering \\
	& \cite{CSR} & CSR   & SR \\
	& \cite{DSIFT} & DSIFT & Transform \\
	& \cite{GFDF} & GFDF  & Filtering \\
	& \cite{GFF} & GFF   & Filtering \\
	& \cite{MSTSR} & MSTSR & Transform+SR \\
	& \cite{MWGF} & MWGF  & Segmentation \bigstrut[b]\\
	\hline
	\multirow{4}[2]{*}{DL based models} & \cite{CNN} & CNN   & Supervised \bigstrut[t]\\
	& \cite{IFCNN} & IFCNN & Unsupervised \\
	& \cite{MFF_NET} & MMFNet & Supervised \\
	& \cite{SESF} & SESF  & Unsupervised \bigstrut[b]\\
	\hline
\end{tabular}%

	\label{tab:method}%
\end{table}%

\begin{table*}[htbp]
	\centering
	\caption{Metrics on MFFW2. The best, the second best and the third best results are highlighted by bold typeface, red and blue, respectively.}
	\resizebox{\textwidth}{!}{
		\begin{tabular}{lccccccccccccc}
			\hline
			& BF    & BRWTSFM & CBF   & CSR   & DSIFT & GFDF  & GFF   & MSTSR & MWGF  & CNN   & IFCNN & MMFNet & SESF \bigstrut\\
			\hline
			MI    & \textcolor[rgb]{ 1,  0,  0}{1.11041 } & 1.04145  & 0.86481  & 0.90291  & \textbf{1.11394 } & \textcolor[rgb]{ .267,  .447,  .769}{1.07927 } & 0.93709  & 0.79136  & 1.05172  & 1.06379  & 0.77777  & 0.91739  & 1.06069  \bigstrut[t]\\
			TE    & \textbf{0.37851 } & 0.36902  & 0.35711  & 0.36007  & \textcolor[rgb]{ 1,  0,  0}{0.37730 } & \textcolor[rgb]{ .267,  .447,  .769}{0.37460 } & 0.36285  & 0.34349  & 0.36948  & 0.37433  & 0.34570  & 0.35683  & 0.36670  \\
			NCIE  & \textcolor[rgb]{ 1,  0,  0}{0.83999 } & 0.83553  & 0.82578  & 0.82669  & \textbf{0.84017 } & \textcolor[rgb]{ .267,  .447,  .769}{0.83772 } & 0.82969  & 0.82192  & 0.83565  & 0.83685  & 0.82105  & 0.82846  & 0.83681  \\
			GBM   & 0.59408  & 0.61646  & 0.52436  & 0.54657  & 0.63253  & \textcolor[rgb]{ .267,  .447,  .769}{0.63469 } & 0.60216  & 0.60297  & \textcolor[rgb]{ 1,  0,  0}{0.65209 } & \textbf{0.66146 } & 0.43159  & 0.42425  & 0.60389  \\
			SF    & -0.06188  & -0.04210  & -0.06579  & -0.05860  & \textcolor[rgb]{ 1,  0,  0}{-0.03101 } & -0.04540  & -0.05152  & \textcolor[rgb]{ .267,  .447,  .769}{-0.03419 } & -0.05450  & -0.05786  & -0.06056  & -0.04734  & \textbf{0.00158 } \\
			SSBM  & \textbf{0.98779 } & 0.94442  & 0.85081  & 0.87093  & 0.93205  & 0.97328  & 0.95175  & 0.91612  & \textcolor[rgb]{ 1,  0,  0}{0.98216 } & \textcolor[rgb]{ .267,  .447,  .769}{0.97580 } & 0.87969  & 0.88438  & 0.95831  \\
			CBM   & \textcolor[rgb]{ 1,  0,  0}{0.74556 } & 0.72685  & 0.65541  & 0.68420  & 0.72038  & \textcolor[rgb]{ .267,  .447,  .769}{0.74257 } & 0.71857  & 0.69597  & \textbf{0.74643 } & 0.73620  & 0.64275  & 0.66439  & 0.73005  \\
			LIF   & 0.38648  & 0.38565  & 0.38885  & 0.38495  & 0.38700  & 0.38603  & 0.38611  & \textcolor[rgb]{ .267,  .447,  .769}{0.38314 } & 0.38634  & 0.38479  & \textcolor[rgb]{ 1,  0,  0}{0.38292 } & \textbf{0.37826 } & 0.38404  \\
			AG    & 3.51687  & 3.56747  & 3.49238  & 3.45226  & \textcolor[rgb]{ .267,  .447,  .769}{3.64229 } & 3.55903  & 3.52980  & \textcolor[rgb]{ 1,  0,  0}{3.64965 } & 3.51950  & 3.51256  & 3.54638  & 3.60238  & \textbf{3.66501 } \\
			MSD   & 0.07902  & 0.07913  & 0.07818  & 0.07895  & 0.07915  & 0.07907  & 0.07904  & \textcolor[rgb]{ 1,  0,  0}{0.08046 } & 0.07907  & \textcolor[rgb]{ .267,  .447,  .769}{0.07989 } & \textbf{0.08054 } & 0.07894  & 0.07954  \\
			GLD   & 17.39289  & 17.63803  & 17.26349  & 17.00870  & \textcolor[rgb]{ .267,  .447,  .769}{17.99025 } & 17.59924  & 17.45074  & \textcolor[rgb]{ 1,  0,  0}{18.03521 } & 17.39076  & 17.36946  & 17.52463  & 17.80669  & \textbf{18.10413 } \bigstrut[b]\\
			\hline
		\end{tabular}%
	}
	\label{tab:MFFW}%
\end{table*}%

\begin{table*}[htbp]
	\centering
	\caption{Metrics on MFFW3. The best, the second best and the third best results are highlighted by bold typeface, red and blue, respectively.}
	\resizebox{\textwidth}{!}{
		\begin{tabular}{lccccccccccccc}
			\hline
			& BF    & BRWTSFM & CBF   & CSR   & DSIFT & GFDF  & GFF   & MSTSR & MWGF  & CNN   & IFCNN & MMFNet & SESF \bigstrut\\
			\hline
			LIF   & 0.36995  & 0.36966  & 0.37290  & \textcolor[rgb]{ .267,  .447,  .769}{0.36772 } & 0.36936  & 0.37157  & 0.37015  & 0.36922  & 0.37250  & 0.36891  & 0.37123  & \textcolor[rgb]{ 1,  0,  0}{0.36768 } & \textbf{0.36616 } \bigstrut[t]\\
			AG    & 4.44186  & 4.49182  & 4.36176  & 4.37234  & \textcolor[rgb]{ .267,  .447,  .769}{4.57004 } & 4.47768  & 4.42958  & \textbf{4.62348 } & 4.41016  & 4.41305  & 4.43954  & \textcolor[rgb]{ 1,  0,  0}{4.61911 } & 4.56945  \\
			MSD   & 0.09115  & 0.09136  & 0.09018  & 0.09124  & 0.09173  & 0.09120  & 0.09119  & \textbf{0.09306 } & 0.09112  & 0.09185  & \textcolor[rgb]{ 1,  0,  0}{0.09286 } & \textcolor[rgb]{ .267,  .447,  .769}{0.09203 } & 0.09178  \\
			GLD   & 22.13581  & 22.37723  & 21.72002  & 21.74381  & \textcolor[rgb]{ .267,  .447,  .769}{22.76412 } & 22.31455  & 22.06844  & \textbf{23.02233 } & 21.95601  & 21.99169  & 22.11054  & \textcolor[rgb]{ 1,  0,  0}{23.00021 } & 22.75713  \bigstrut[b]\\
			\hline
		\end{tabular}%
		\label{tab:MFFW3}%
	}
\end{table*}%

\subsection{Metrics}
Since there is no ground-truth image for MFF task, the no-reference image assessment metrics are widely employed in this case. Eleven metrics are used in our paper. In what follows, $A,B$ denote the source images, and $F$ is the fusion image. $M,N$ are the height and width of an image, respectively.

\subsubsection{Information theory based metric}
Mutual information (MI) is a terminology in information theory defined by 
\begin{equation*}
	{\rm MI}_{XY}= \iint f_{XY}(x,y)\log_2 \frac{f_{XY}(x,y)}{f_X(x)f_Y(y)} dxdy,
\end{equation*}
where $X,Y$ denote two random variables, $f_X,f_Y$ are corresponding probability density function (PDF) and $f_{XY}$ is their joint PDF. MI reflects the correlation between $X$ and $Y$. In the context of MFF, PDF is replaced with the normalized histogram of pixels, and ${\rm Q_{MI}}$ is defined by \cite{MI}
\begin{equation*}
	Q_{MI} = \frac{1}{2} ({\rm MI}_{AF}+{\rm MI}_{BF}).
\end{equation*}  
Except MI, there are other metrics that are used to assess image quality in image fusion field, such as Tsallis entropy based metric ($Q_{TE}$) \cite{TE} and nonlinear correlation information entropy ($Q_{NCIE}$) \cite{NCIE}. The principle of these metrics is to measure the agreement between fused and source images and to overcome the drawback of MI (i.e., sensitivity to noise).
\subsubsection{Gradient-based metric}
Gradient-based metric (GBM) \cite{xydeas2000objective} measures how much gradient information transferred to the fusion image. It is defined by 
\begin{equation*}
	Q_{GBM} = \frac{\sum_{i=1}^{M}\sum_{j=1}^{N} m^{AF}_{ij}w^A_{ij}+m^{BF}_{ij}w^B_{ij}  }
	{\sum_{i=1}^{M}\sum_{j=1}^{N} w^A_{ij}+w^B_{ij} },
\end{equation*}
where 
\begin{equation*}
	m^{XF}_{ij}= e^{XF}_{ij}o^{XF}_{ij}, (X=A,B).
\end{equation*}
Here, $e^{XF}_{ij}$ and $o^{XF}_{ij}$ are the magnitude of the edge strength and orientation preservation of $(i,j)^{\rm th}$ pixel, respectively. And, $w^X_{ij}$ is the gradient strength of the source image $X$.

\subsubsection{Structure similarity-based metric}
Structure similarity-based metric (SSBM) is an extension of structure similarity $s$ for image fusion task \cite{SSBM}. Note that give two images $X$ and $Y$, $s_{XY}$ evaluates their similarity window by window. For a local window, SSBM is defined by
\begin{equation*}
Q_{SSBM}=
\begin{cases}
w s_{AF}+(1-w)s_{BF}, & {\rm if \ } s_{AB}\ge 0.75, \\
\max\{s_{AF}, s_{BF} \}, & {\rm otherwise},
\end{cases}
\end{equation*}
where weight is $w=D_A/(D_A+D_B)$, and $D_X$ is the variance in this local window.

\subsubsection{Spatial frequency based metric}
Zheng et al. \cite{SF} used four first-order gradients with different directions to define spatial frequency as follows
$$
{\rm SF} = \sqrt{{\rm HF}+{\rm VF}+{\rm MDF}+{\rm SDF}}.
$$ 
For a pixel in an image $X$, there are
$$
{\rm HF} = \frac{\sum_{i=1}^{M}\sum_{j=1}^{N}(\nabla_h X)^2_{ij}}{MN},
{\rm MDF} = \frac{\sum_{i=1}^{M}\sum_{j=1}^{N}(\nabla_{md} X)^2_{ij}}{MN\sqrt{2}},
$$
$$
{\rm VF} = \frac{\sum_{i=1}^{M}\sum_{j=1}^{N}(\nabla_v X)^2_{ij}}{MN},
{\rm SDF} = \frac{\sum_{i=1}^{M}\sum_{j=1}^{N}(\nabla_{sd} X)^2_{ij}}{MN\sqrt{2}}.
$$
Four gradient operators $\nabla_g (g=h,v,md,sd)$ are defined by 
$$
\nabla_h=X_{ij}-X_{i,j-1},\nabla_md=X_{ij}-X_{i-1,j-1},
$$
$$
\nabla_v=X_{ij}-X_{i-1,j},\nabla_sd=X_{ij}-X_{i-1,j+1},
$$
With source images $A$ and $B$, they defined the reference gradients
$$
\nabla_g R_{ij} = \max\{|\nabla_g A_{ij}|,|\nabla_g B_{ij}|\}.
$$ 
Thereafter, it is able to obtain reference spatial frequency (denoted by ${\rm SF}_R$) and spatial frequency of fused image (denoted by ${\rm SF}_F$). At last, Zheng's metric is given by the relative ratio of SF error ($Q_{SF}$), namely,
$$
Q_{SF}=({\rm SF}_F-{\rm SF}_R)/{\rm SF}_R.
$$
\subsubsection{Chen-blum metric}
Chen-blum metric (CBM) \cite{CB} compares the contrast features to source images with a fusion image. It is defined by 
\begin{equation*}
	Q_{CBM}=\frac{1}{MN} \sum_{i=1}^{M}\sum_{j=1}^{N} \lambda^A_{ij} c^{AF}_{ij}+\lambda^B_{ij}c^{BF}_{ij},
\end{equation*}
where $\lambda^X$ is the saliency map and $c^{XF}$ is the contrast preserved in the fusion image from source image $ X $.

\subsubsection{Linear index of fuzziness}
Linear index of fuzziness (LIF) \cite{LIF} is used to evaluate the enhancement of fusion images, defined by
\begin{equation*}
	Q_{LIF} = \frac{2}{MN}\sum_{i=1}^{M}\sum_{j=1}^{N} \min \{p_{ij}, (1-p_{ij})\},
\end{equation*}
where 
\begin{equation*}
	p_{ij} = \sin\left(\frac{\pi}{2}(1-F_{ij}/F_{\max})\right).
\end{equation*}

\subsubsection{Average gradient}
Average gradient (AG) \cite{AG} measures the gradient information of fusion images, defined by
\begin{equation*}
	Q_{AG}=\frac{1}{MN}\sum_{i=1}^{M}\sum_{j=1}^{N}\frac{1}{4}\sqrt{(\nabla_h F)^2_{ij}+(\nabla_v F)^2_{ij}},
\end{equation*}
where $\nabla_h$ and $\nabla_v$ denote the horizontal and vertical gradient operators, respectively.

\subsubsection{Mean square deviation}
Mean square deviation (MSD) \cite{MFF_NET} measures image detail richness, defined by
\begin{equation*}
	Q_{MSD}=\frac{1}{MN}\sqrt{\sum_{i=1}^{M}\sum_{j=1}^{N} (F_{ij}-\bar{F})^2},
\end{equation*}
where $\bar{F}$ denotes the mean pixel value of fusion image.

\subsubsection{Gray level difference}
Gray level difference (GLD) \cite{MFF_NET} is another gradient-based metric, defined by
\begin{equation*}
	Q_{GLD} = \frac{\sum_{i=1}^{M}\sum_{j=1}^{N} |\nabla_h F|+|\nabla_v F|}{MN}.
\end{equation*}

\textbf{Remark}: Fusion images are better if all metrics are larger except for LIF. In spite of the generalized usage of the first seven metrics (i.e., MI, TE, NCIE, GBM, SF, SSBM and CBM), they are criticized for computing agreements of the fusion image with source images \cite{Liu2012,pena2019multiple}. In addition, they do not work if the number of source images is larger than two. Some investigations have pointed out that they often lead to confused judgment \cite{Liu2012}. As for the last four metrics (i.e., LIF, AG, MSD and GLD), they measure the features of fusion images and there is no business of source images. Generally speaking, they are more reliable in terms of whether the boundaries are clear \cite{MFF_NET}. However, they will lose efficiency if there are artifacts in fusion images.

\begin{figure*}
	\centering
	\subfigure[BF]       {\includegraphics[width=0.13\linewidth]{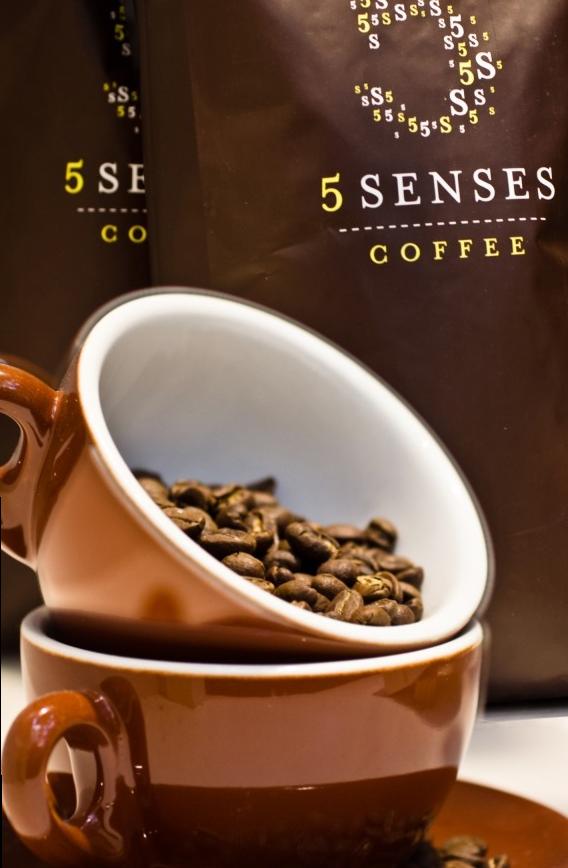}}
	\subfigure[BRW]      {\includegraphics[width=0.13\linewidth]{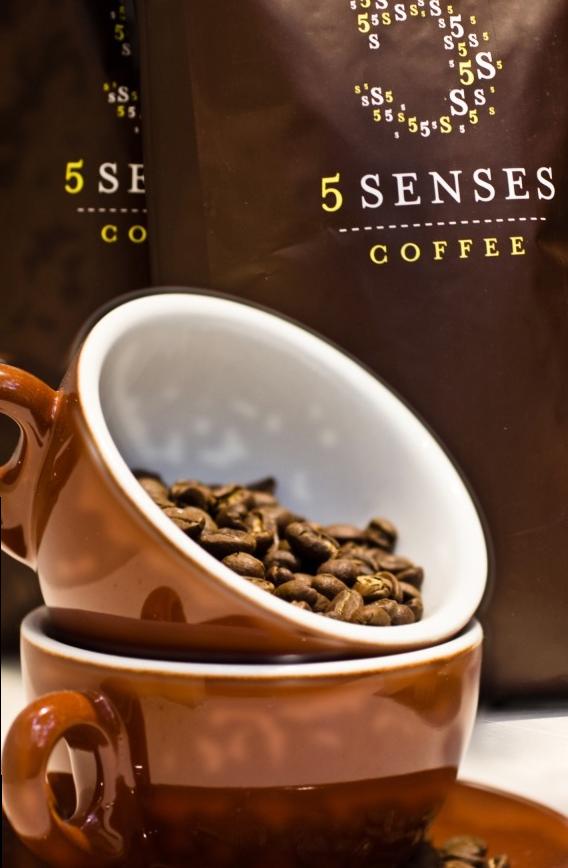}}
	\subfigure[CBF]      {\includegraphics[width=0.13\linewidth]{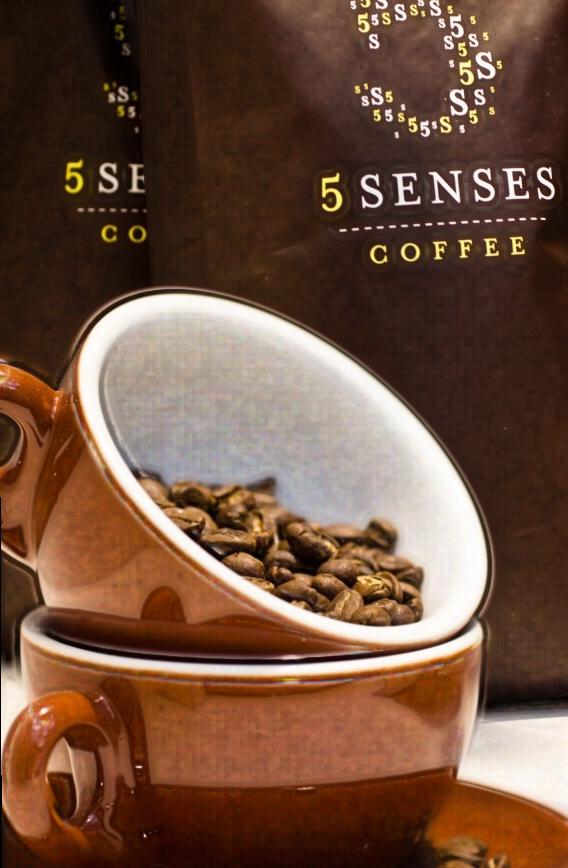}}
	\subfigure[CNN]      {\includegraphics[width=0.13\linewidth]{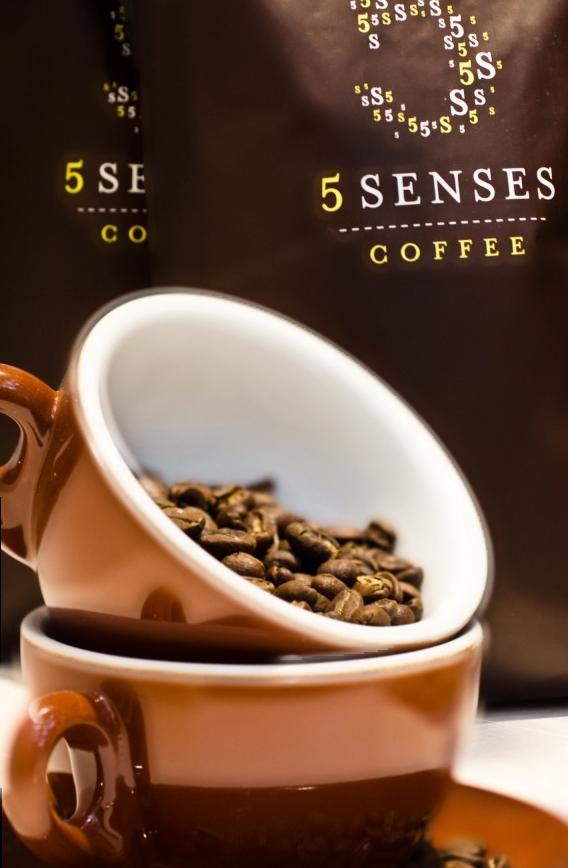}}
	\subfigure[CSR]      {\includegraphics[width=0.13\linewidth]{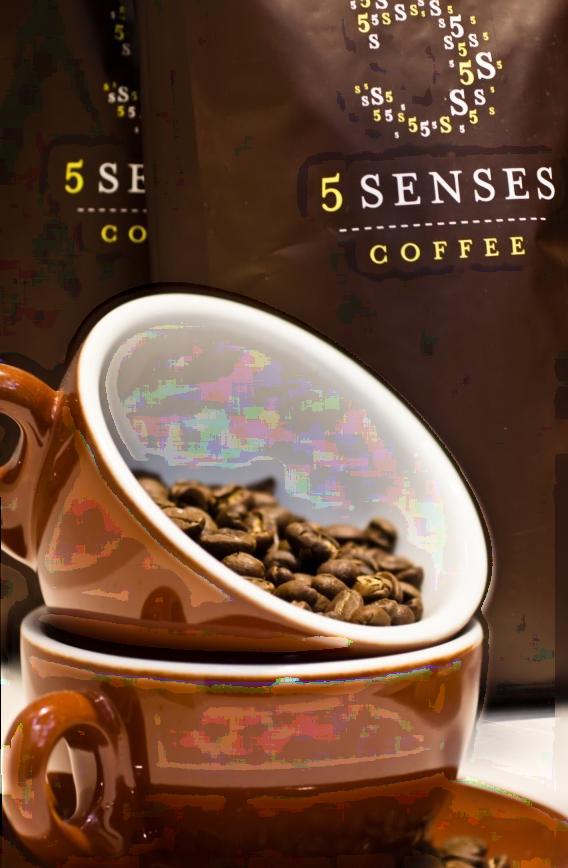}}
	\subfigure[DSIFT]    {\includegraphics[width=0.13\linewidth]{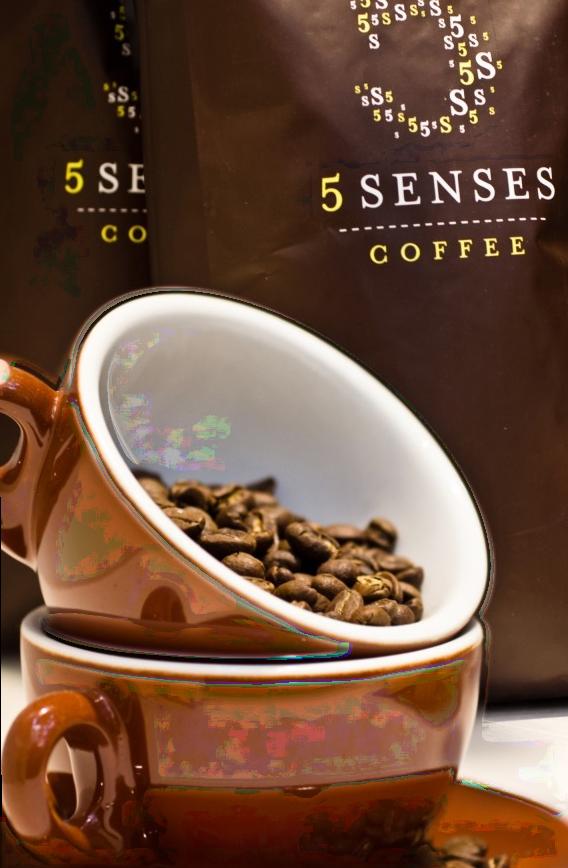}}
	\subfigure[GFDF]     {\includegraphics[width=0.13\linewidth]{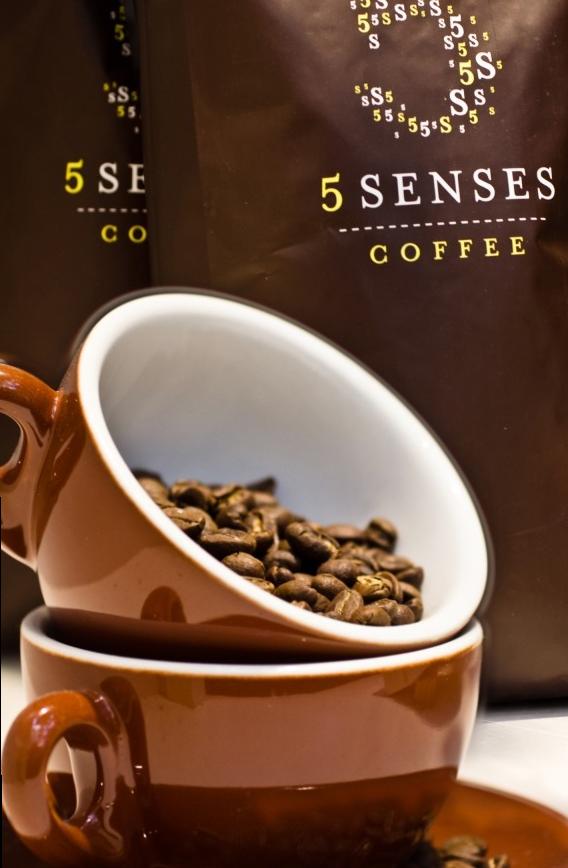}}
	\subfigure[GFF]      {\includegraphics[width=0.13\linewidth]{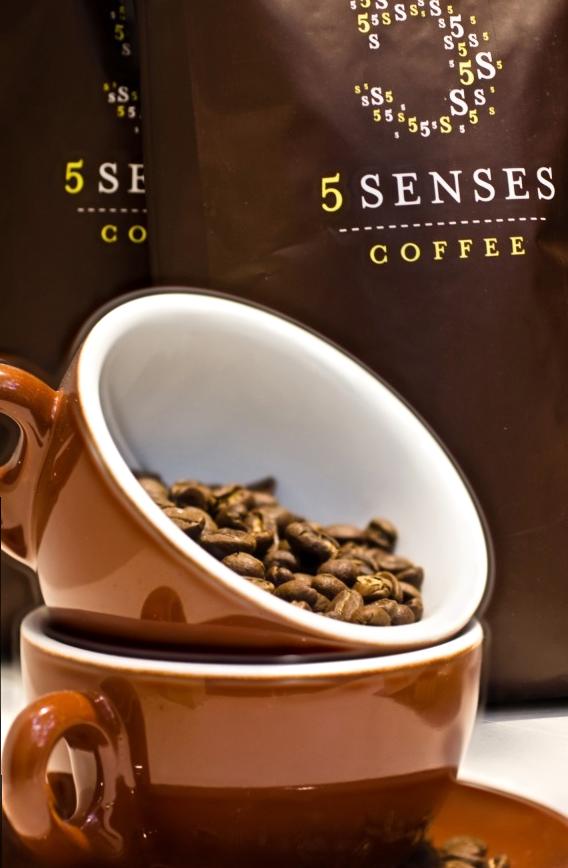}}
	\subfigure[IFCNN]    {\includegraphics[width=0.13\linewidth]{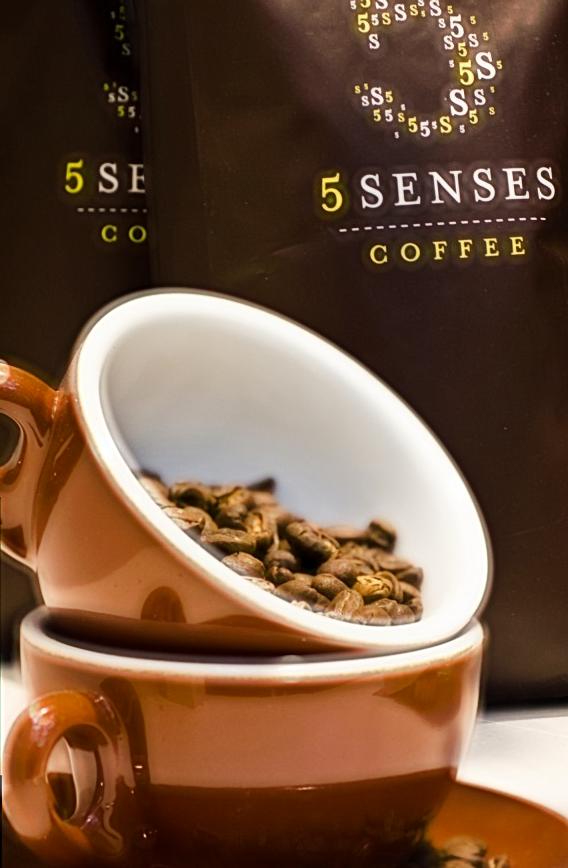}}
	\subfigure[MMFNet]   {\includegraphics[width=0.13\linewidth]{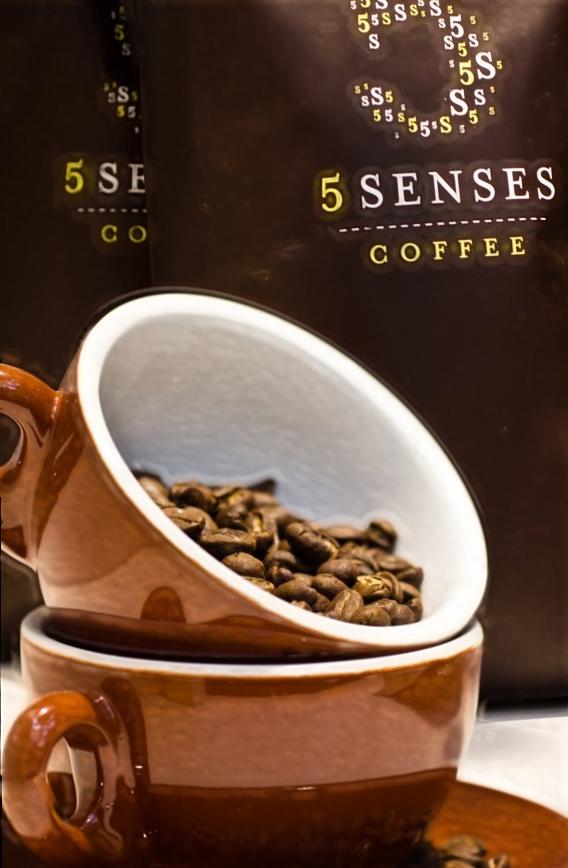}}
	\subfigure[MSTSR]    {\includegraphics[width=0.13\linewidth]{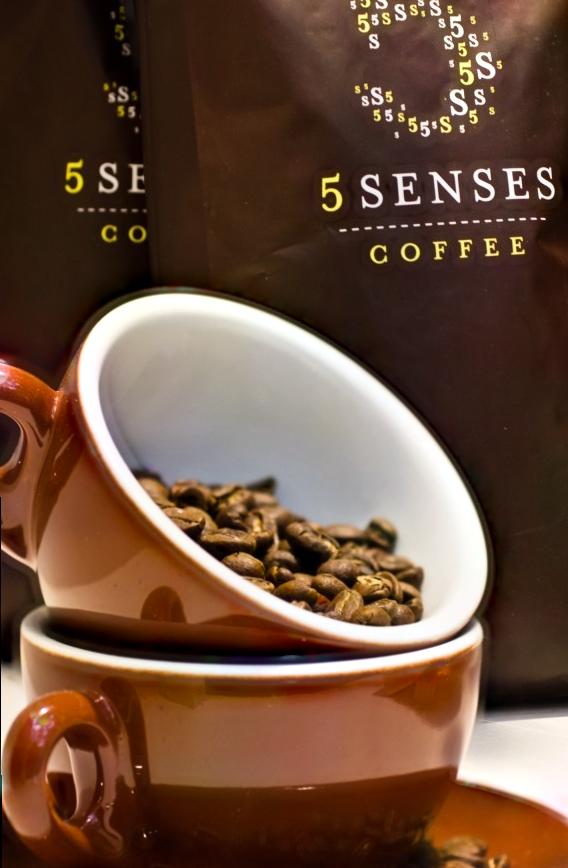}}
	\subfigure[MWGF]     {\includegraphics[width=0.13\linewidth]{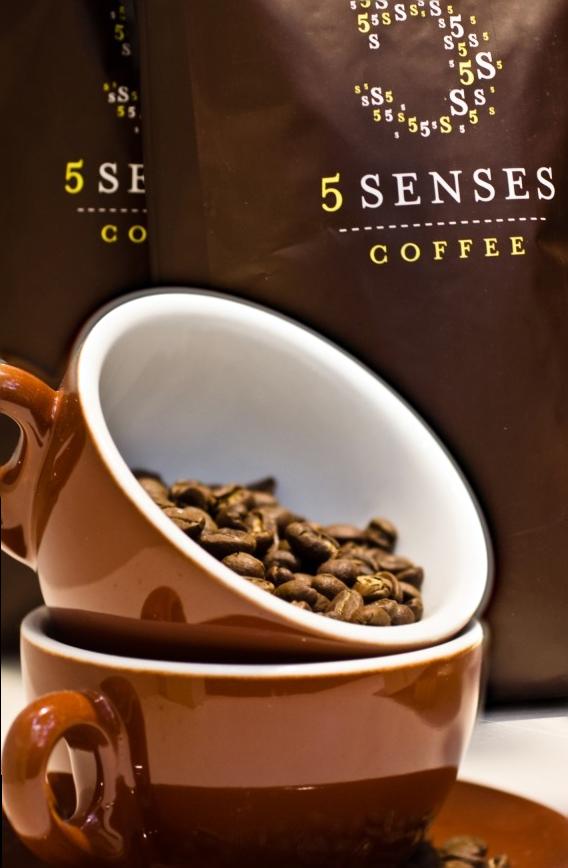}}
	\subfigure[SESF]     {\includegraphics[width=0.13\linewidth]{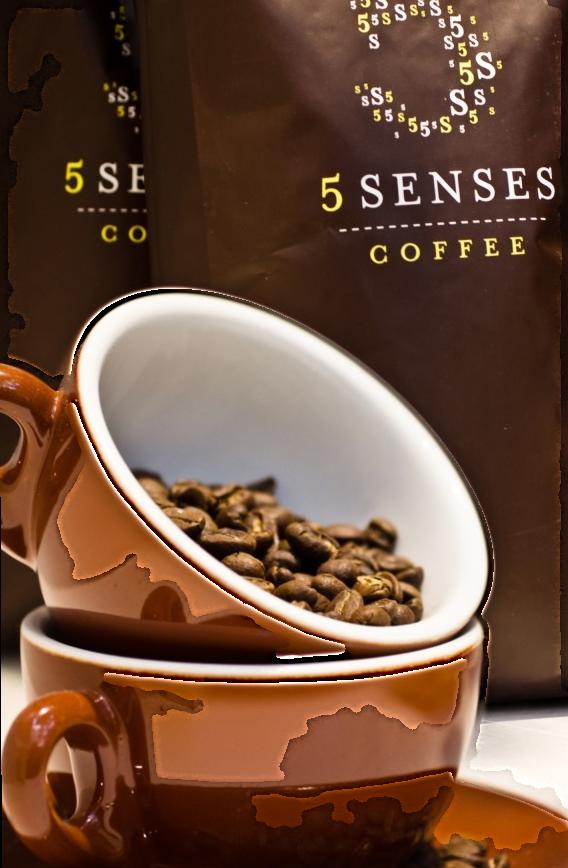}}
	\subfigure[Reference]{\includegraphics[width=0.13\linewidth]{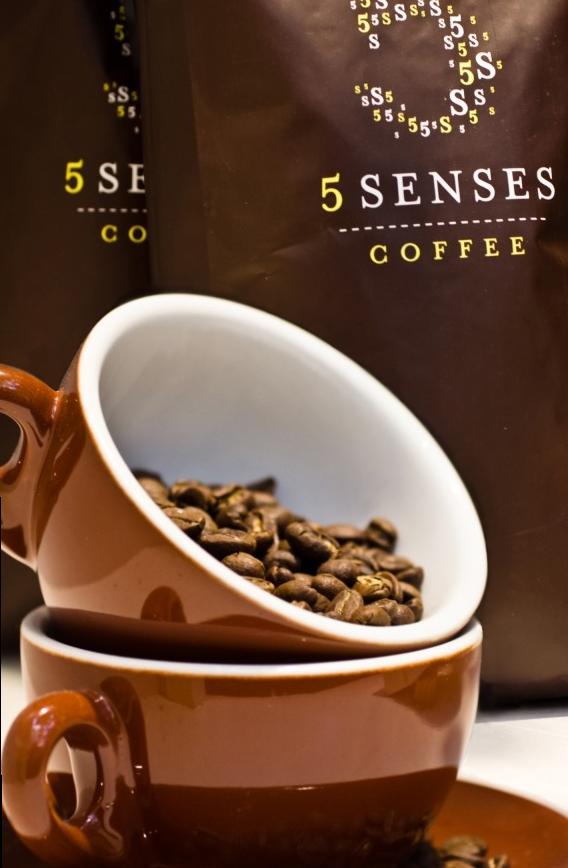}}
	\caption{The fused images of No. 6 pair in MFFW2 dataset.}
	\label{fig:coffee}
\end{figure*}
\begin{figure*}
	\centering
	\subfigure[BF]       {\includegraphics[width=0.13\linewidth]{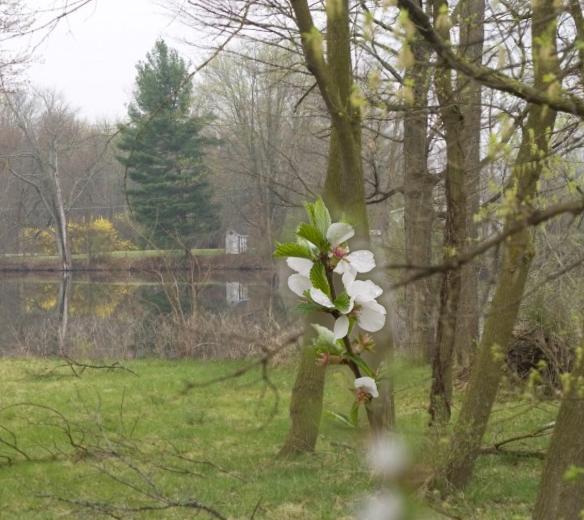}}
	\subfigure[BRW]      {\includegraphics[width=0.13\linewidth]{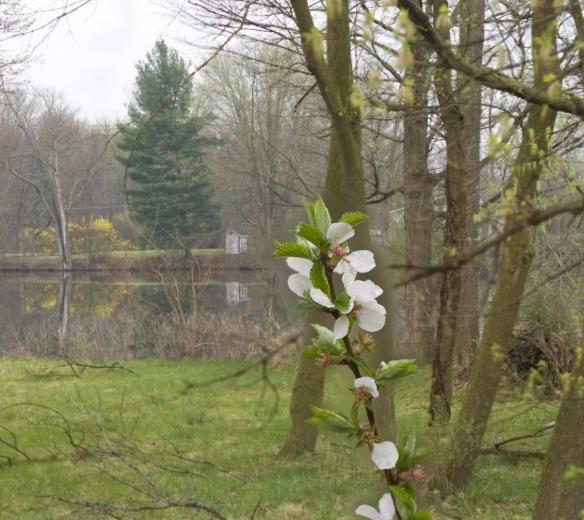}}
	\subfigure[CBF]      {\includegraphics[width=0.13\linewidth]{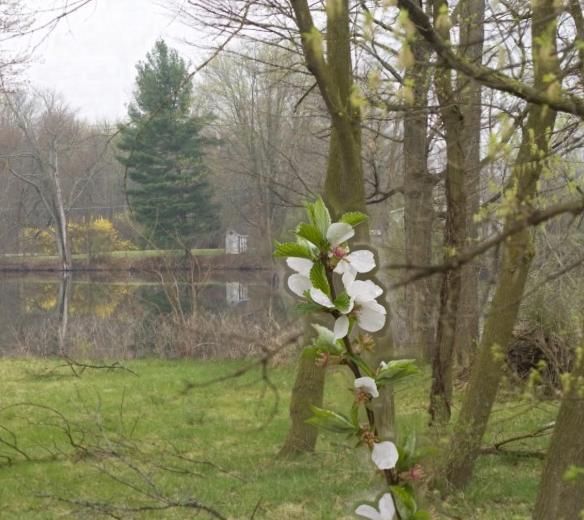}}
	\subfigure[CNN]      {\includegraphics[width=0.13\linewidth]{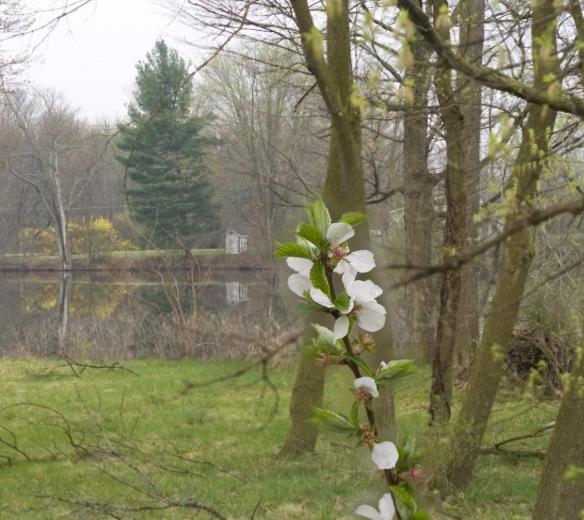}}
	\subfigure[CSR]      {\includegraphics[width=0.13\linewidth]{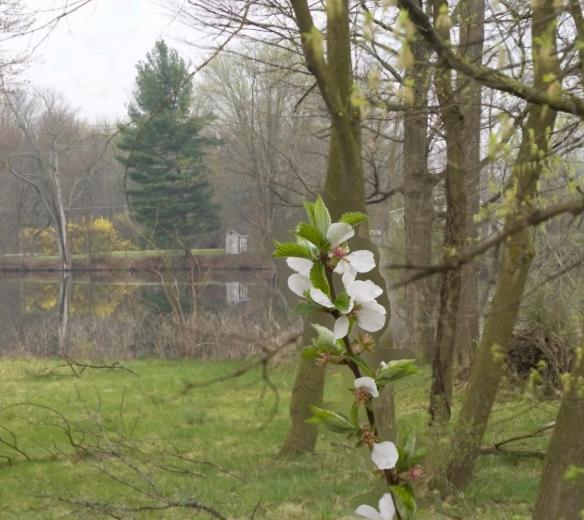}}
	\subfigure[DSIFT]    {\includegraphics[width=0.13\linewidth]{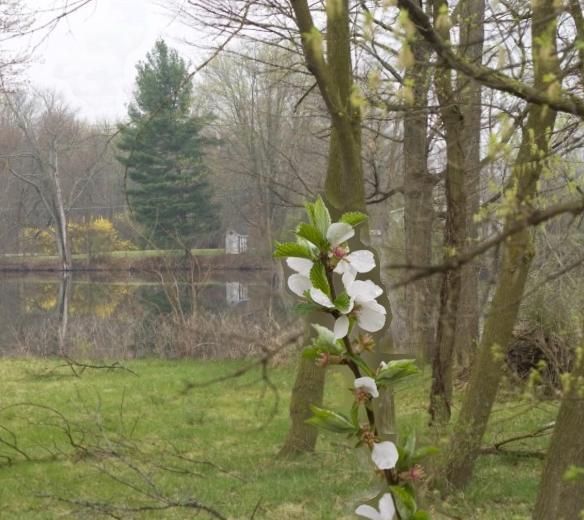}}
	\subfigure[GFDF]     {\includegraphics[width=0.13\linewidth]{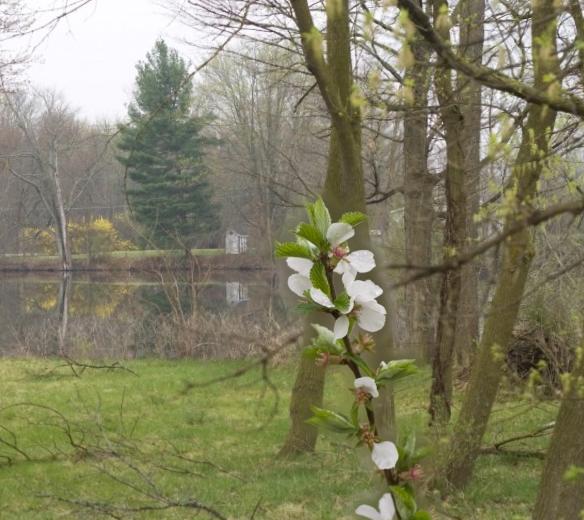}}
	\subfigure[GFF]      {\includegraphics[width=0.13\linewidth]{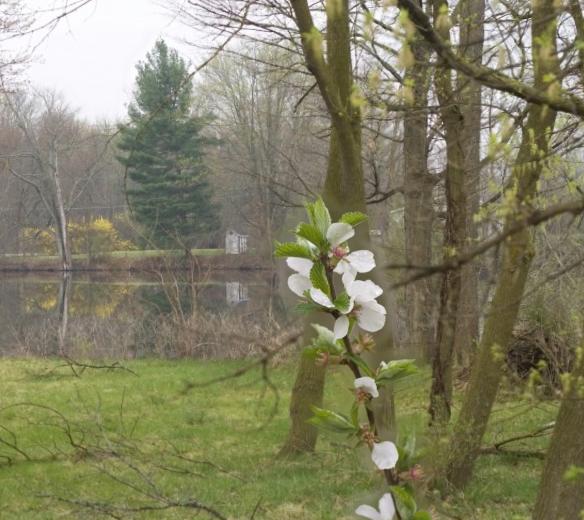}}
	\subfigure[IFCNN]    {\includegraphics[width=0.13\linewidth]{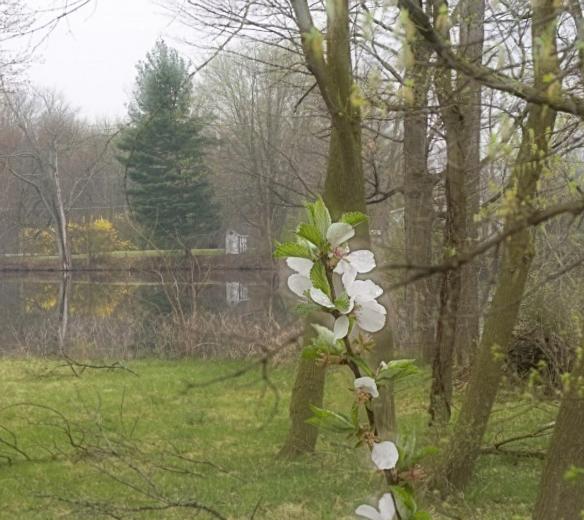}}
	\subfigure[MMFNet]   {\includegraphics[width=0.13\linewidth]{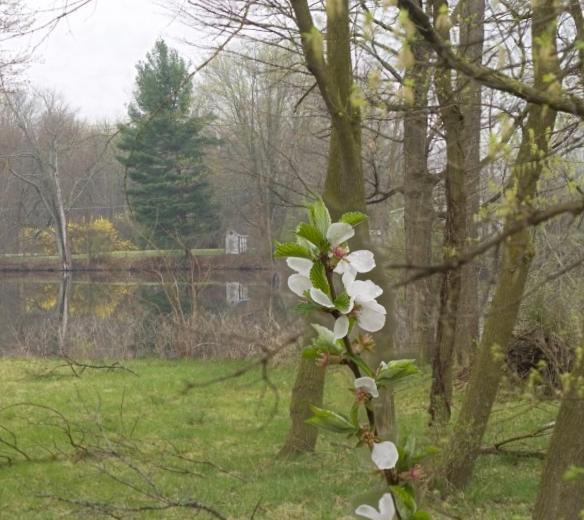}}
	\subfigure[MSTSR]    {\includegraphics[width=0.13\linewidth]{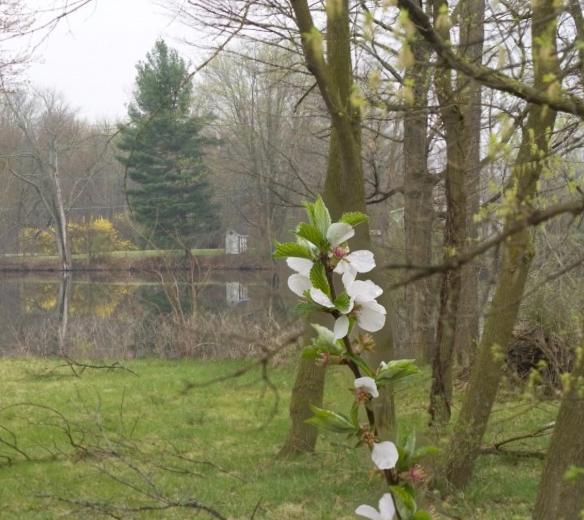}}
	\subfigure[MWGF]     {\includegraphics[width=0.13\linewidth]{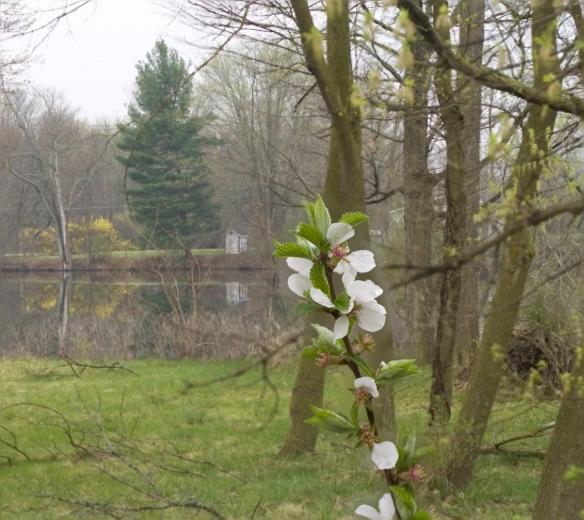}}
	\subfigure[SESF]     {\includegraphics[width=0.13\linewidth]{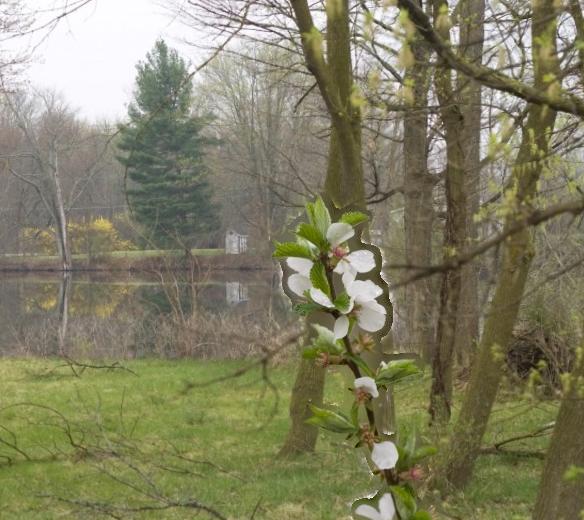}}
	\subfigure[Reference]{\includegraphics[width=0.13\linewidth]{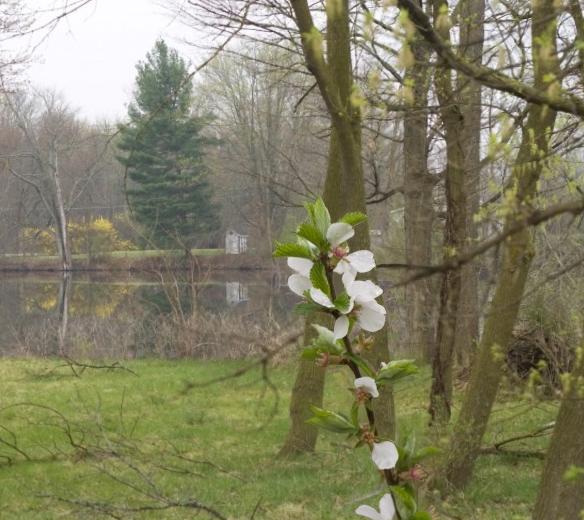}}
	\caption{The fused images of No. 7 pair in MFFW2 dataset.}
	\label{fig:flower}
\end{figure*}
\subsection{Results}
\subsubsection{Metric comparison}
Tables \ref{tab:MFFW} and \ref{tab:MFFW3} list the evaluation metrics on MFFW2 and MFFW3, respectively. For each pair on MFFW3 dataset, the number of source images are larger than two. As a result, we only report the values of LIF, AG, MSD and GLD on MFFW3. 

On MFFW2 dataset, considering the first seven metrics, the best four methods are BF, DSIFT, GFDF and MWGF. In terms of the last four metrics, the best four methods are MSTSR, SESF, IFCNN and DSIFT. On MFFW3 dataset, the best four methods are MSTSR, MMFNet, DSIFT and SESF. Hence, DL based models have no significant superiority on our datasets. 
\begin{figure*}[htbp]
	\centering
	\subfigure[BF]       {\includegraphics[width=0.13\linewidth]{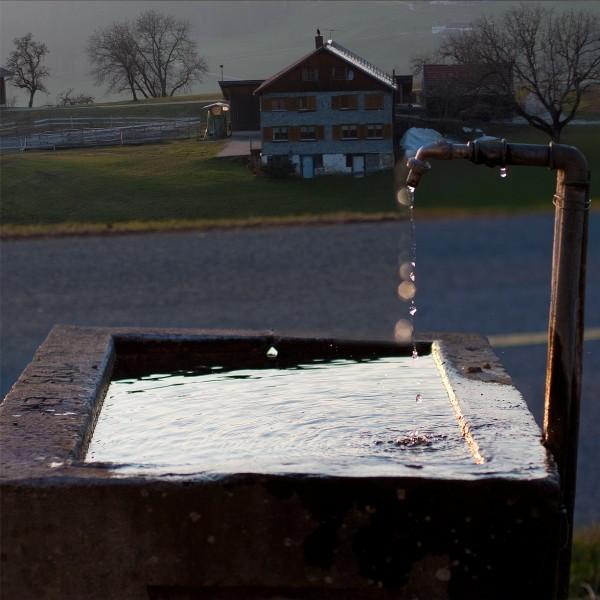}}
	\subfigure[BRW]      {\includegraphics[width=0.13\linewidth]{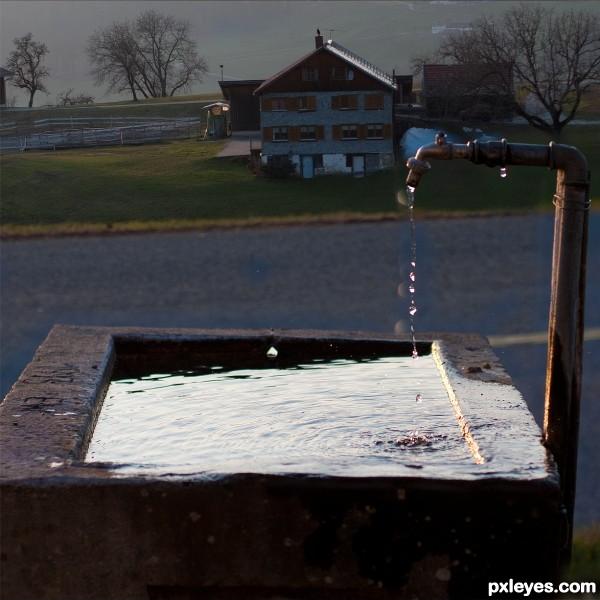}}
	\subfigure[CBF]      {\includegraphics[width=0.13\linewidth]{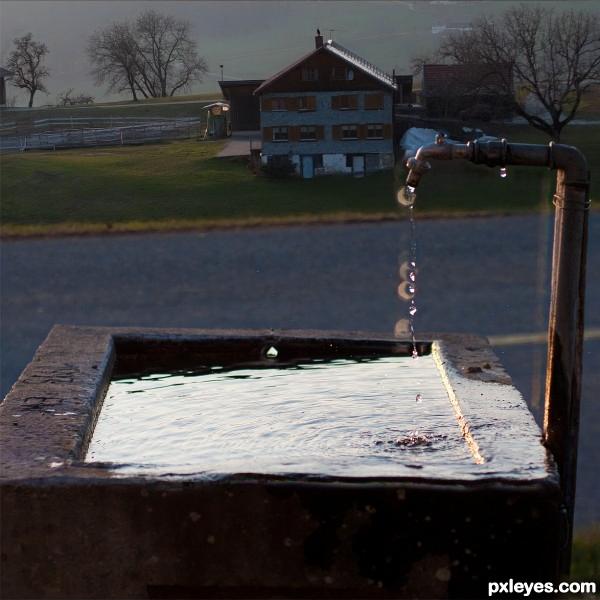}}
	\subfigure[CNN]      {\includegraphics[width=0.13\linewidth]{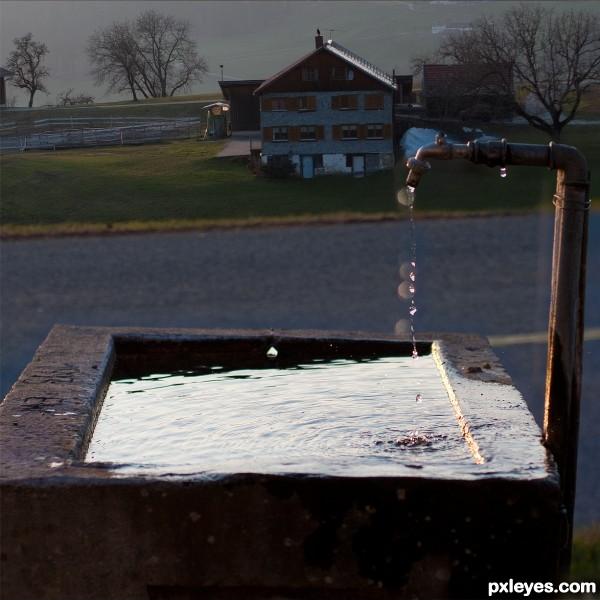}}
	\subfigure[CSR]      {\includegraphics[width=0.13\linewidth]{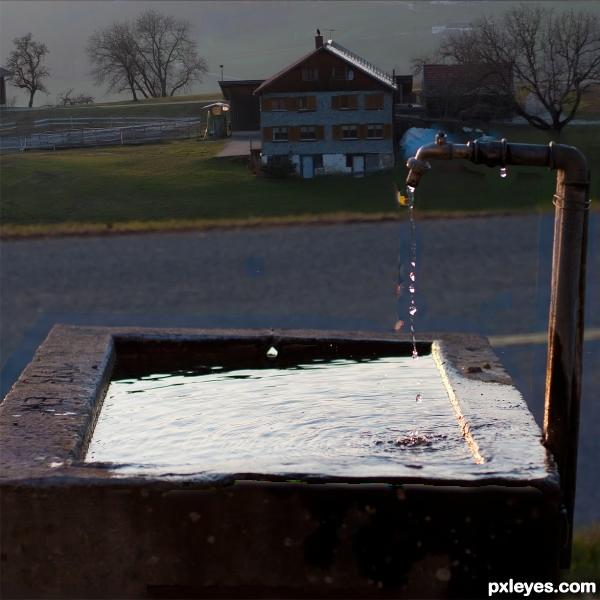}}
	\subfigure[DSIFT]    {\includegraphics[width=0.13\linewidth]{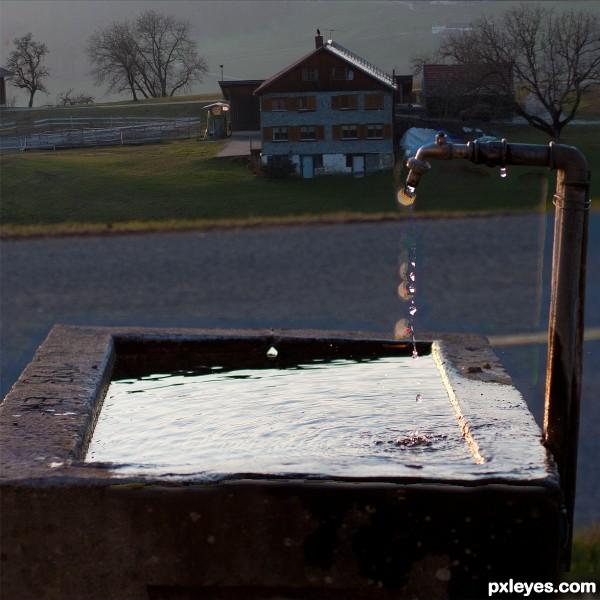}}
	\subfigure[GFDF]     {\includegraphics[width=0.13\linewidth]{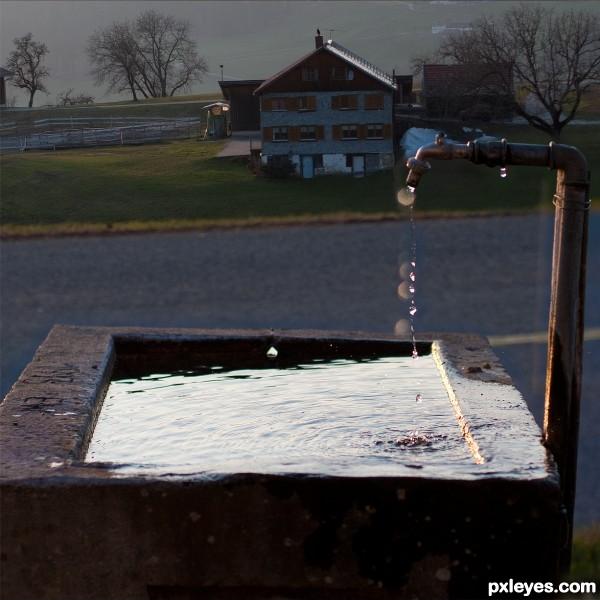}}
	\subfigure[GFF]      {\includegraphics[width=0.13\linewidth]{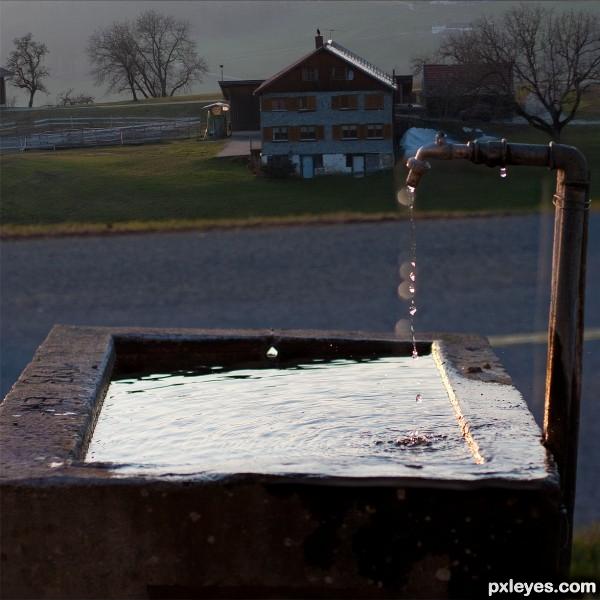}}
	\subfigure[IFCNN]    {\includegraphics[width=0.13\linewidth]{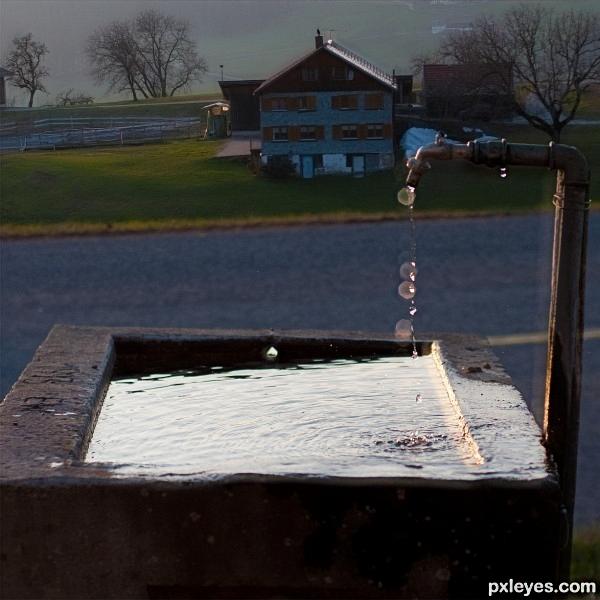}}
	\subfigure[MMFNet]   {\includegraphics[width=0.13\linewidth]{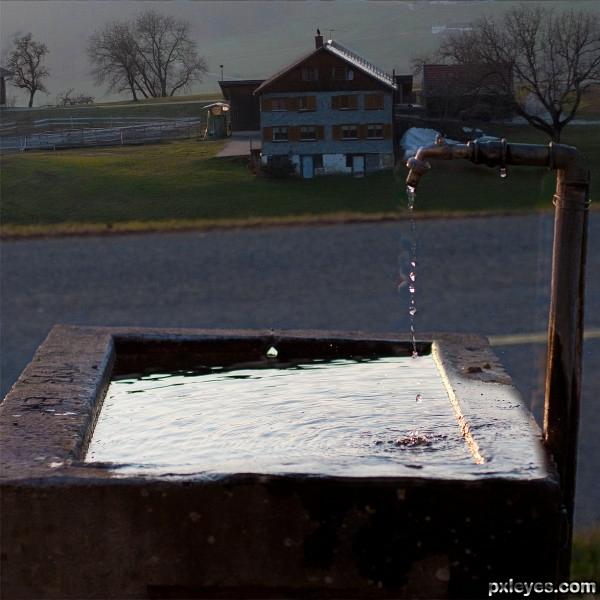}}
	\subfigure[MSTSR]    {\includegraphics[width=0.13\linewidth]{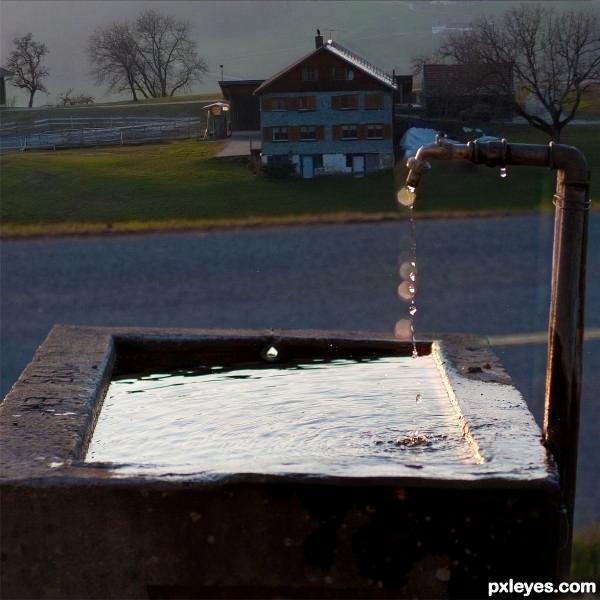}}
	\subfigure[MWGF]     {\includegraphics[width=0.13\linewidth]{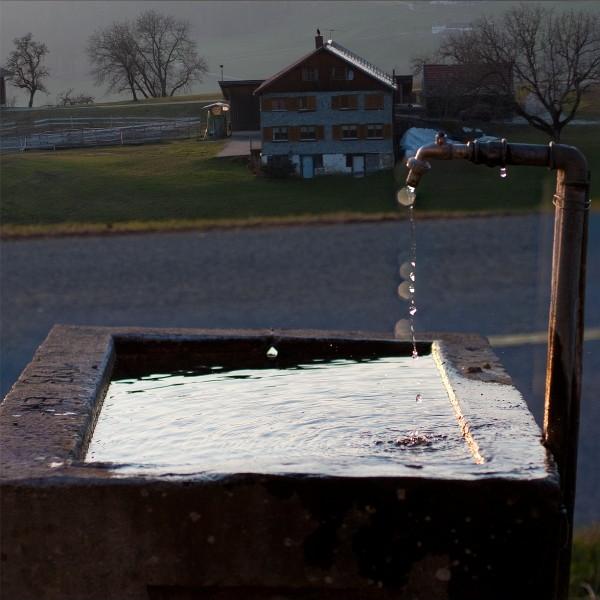}}
	\subfigure[SESF]     {\includegraphics[width=0.13\linewidth]{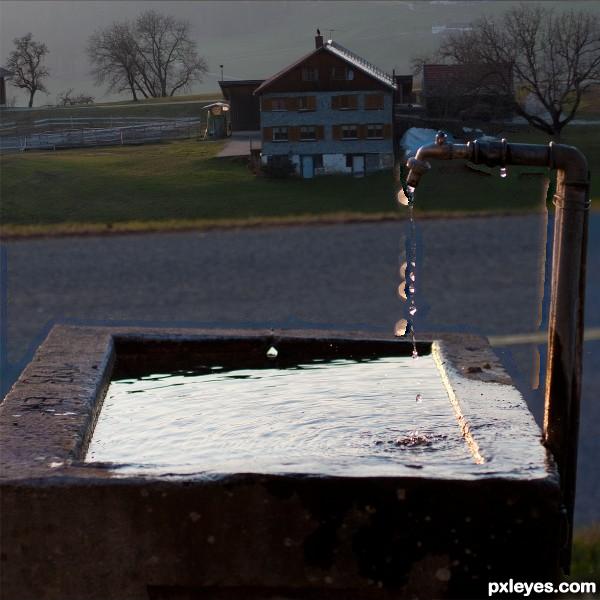}}
	\subfigure[Reference]{\includegraphics[width=0.13\linewidth]{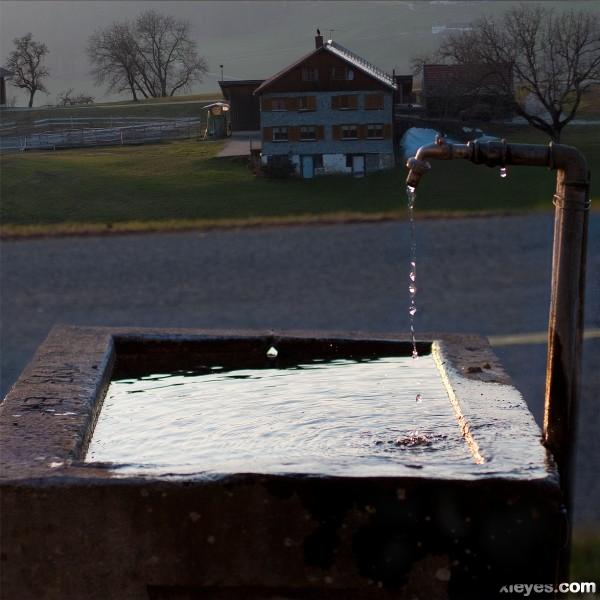}}
	\caption{The fused images of No. 10 pair in MFFW2 dataset.}
	\label{fig:pipe}
\end{figure*}

\begin{figure*}[htbp]
	\centering
	\subfigure[BF]       {\includegraphics[width=0.13\linewidth]{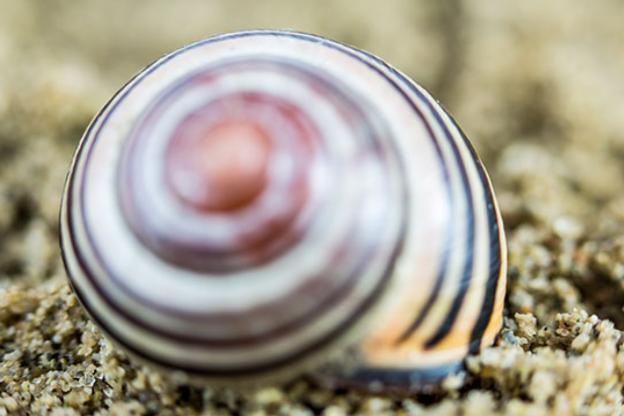}}
	\subfigure[BRW]      {\includegraphics[width=0.13\linewidth]{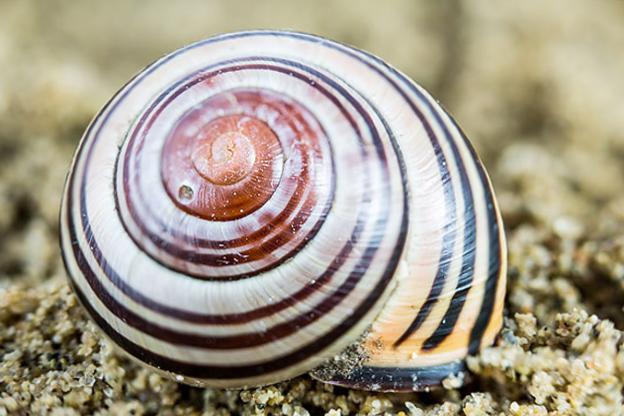}}
	\subfigure[CBF]      {\includegraphics[width=0.13\linewidth]{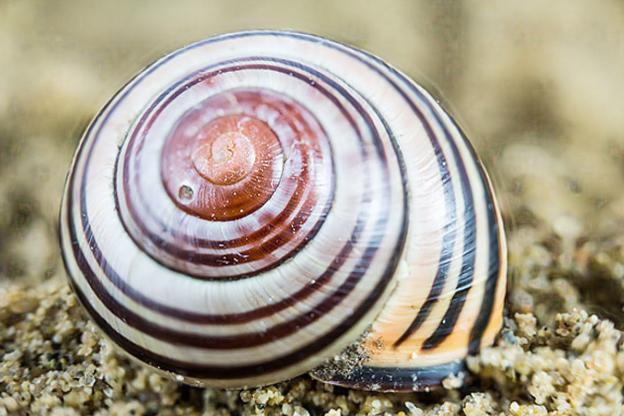}}
	\subfigure[CNN]      {\includegraphics[width=0.13\linewidth]{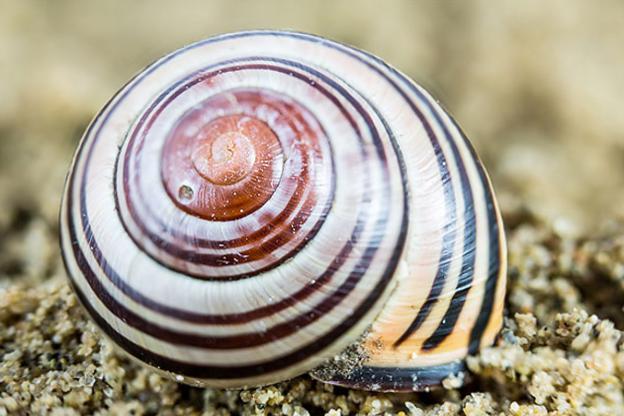}}
	\subfigure[CSR]      {\includegraphics[width=0.13\linewidth]{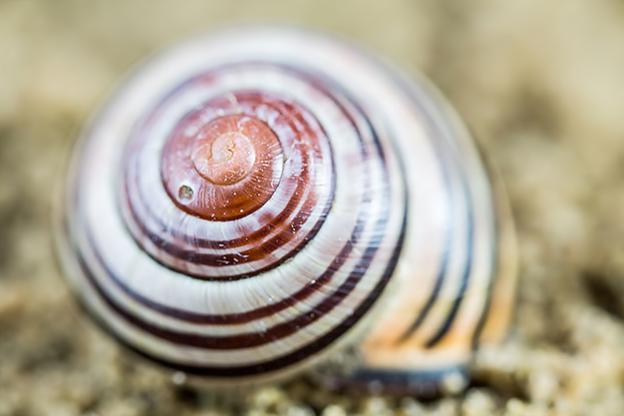}}
	\subfigure[DSIFT]    {\includegraphics[width=0.13\linewidth]{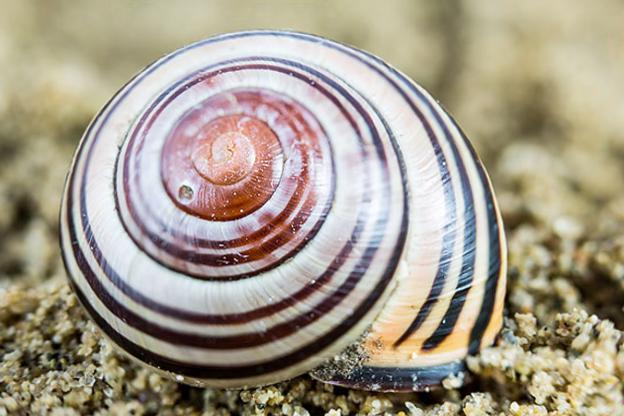}}
	\subfigure[GFDF]     {\includegraphics[width=0.13\linewidth]{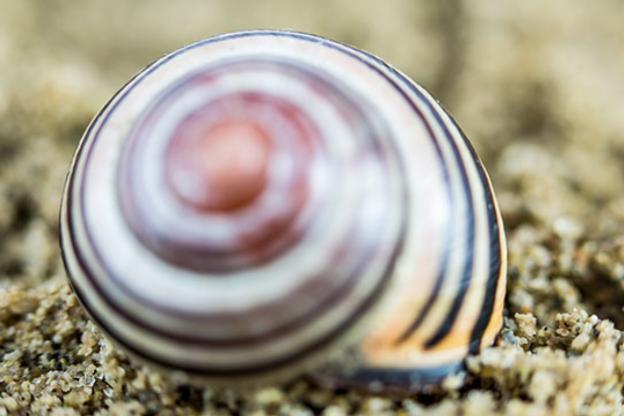}}
	\subfigure[GFF]      {\includegraphics[width=0.13\linewidth]{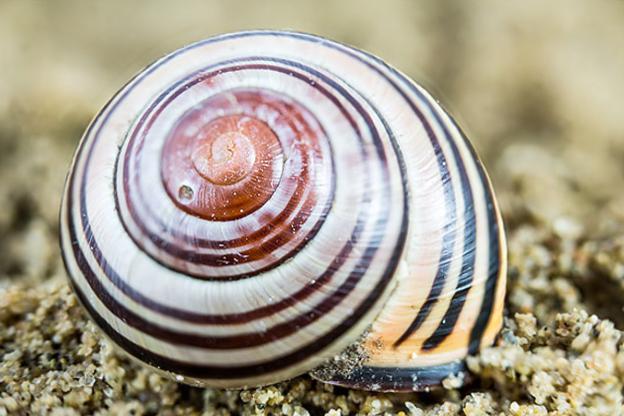}}
	\subfigure[IFCNN]    {\includegraphics[width=0.13\linewidth]{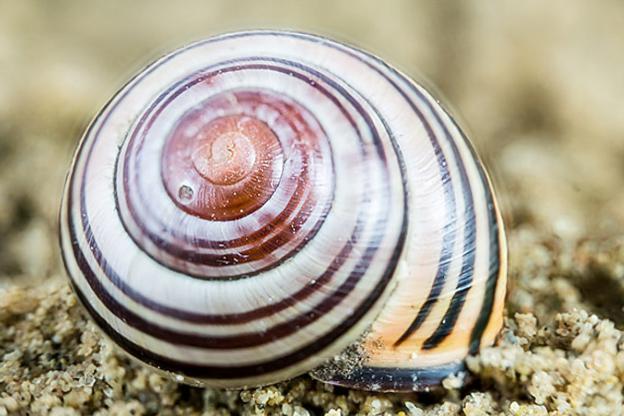}}
	\subfigure[MMFNet]   {\includegraphics[width=0.13\linewidth]{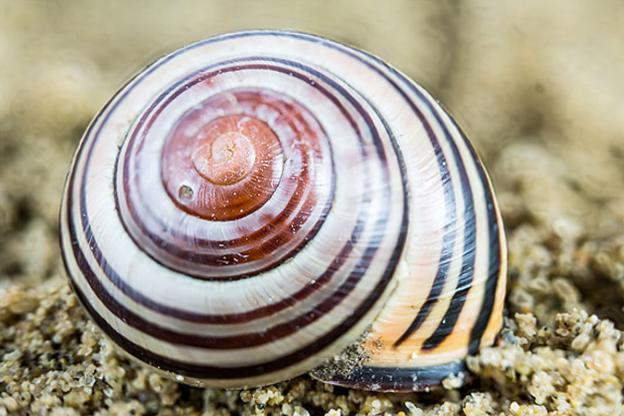}}
	\subfigure[MSTSR]    {\includegraphics[width=0.13\linewidth]{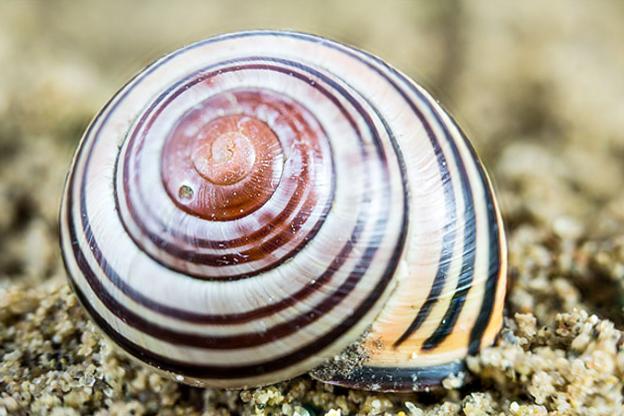}}
	\subfigure[MWGF]     {\includegraphics[width=0.13\linewidth]{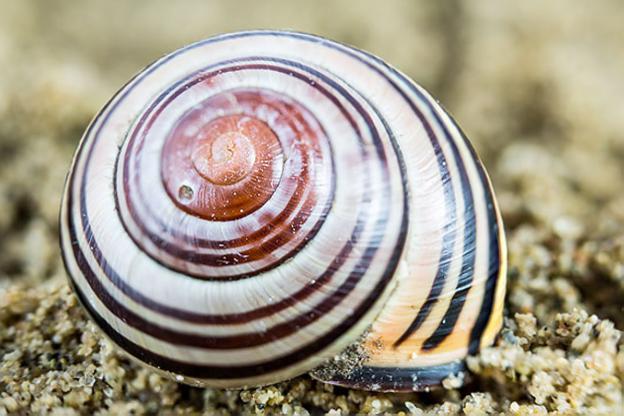}}
	\subfigure[SESF]     {\includegraphics[width=0.13\linewidth]{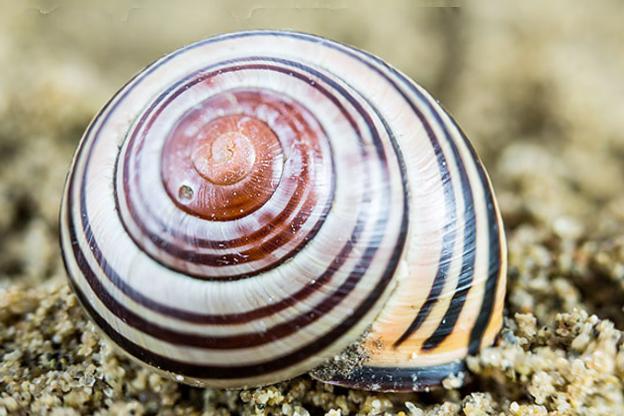}}
	\caption{The fused images of No. 4 pair in MFFW3 dataset.}
	\label{fig:shell}
\end{figure*}
\subsubsection{Visual comparison}
In order to obtain more insights, we would like to pay attention to visual comparison. The main purpose is to check whether SOTA algorithms can robustly deal with DSE in different scenes. Five representative pairs of images are displayed and analyzed in the next.

Fig. \ref{fig:coffee} shows fused images of No. 6 pair of MFFW2 dataset. There are two cups in foreground and two coffee bags in background. The difficulty of this pair is the strong DSE which occurs in the edges of cups and characters printed on coffee bags. Compared with the reference image which is manually edited, it is observed that no method can match up to our expectations. First of all, there is color distortion in the images fused by CBF, CSR and DSIFT, since they process RGB images channel by channel. Secondly, there are visible artifacts in the images produced by BRW, MMFNet and SESF. The reason is that focus maps are mistakenly detected. Thirdly, significant DSE can be observed in the images merged by other methods. 

Fig. \ref{fig:flower} shows fused images of No. 7 pair of MFFW2 dataset. In background, there are many trees, a lawn and a river. In foreground, there is a flower. There is the DSE in the boundary of focus map. On account of the DSE, there is a area around the flower, which is blurred in both source images. It is found that all methods breaks down for this pair. Specifically, the white artifacts exist two tree behind the flower.

Fig. \ref{fig:pipe} shows fused images of No. 10 pair of MFFW2 dataset. A waterpipe and a pool are in foreground, and a cabin is in background. It is worth pointing out that the waterpipe and drops expand when they are not in focus. It is found that BRW and MMFNet perform well in this case. In contrast, both blurred and clear drops exist in the fused images generated by other methods. 

Fig. \ref{fig:shell} shows fused images of No. 4 pair of MFFW3 dataset. There is a shell lying in a beach. The focuses of source images are at different parts of this shell. Note that we do not provide reference images for MFFW3 dataset owing to the extremely high difficulty and complexity. As shown in Fig. \ref{fig:shell}, BF, CSR and GFDF evidently provide blurry images. By and large, other methods generate images with clear profiles, while their performance varies in the edge of this shell. As displayed in Fig. \ref{fig:image3}, the focus point of the first source image is payed in the center of this shell. Consequently, the edge of this shell expands on account of the DSE. It is found that all methods except DSIFT and MWGF generate obvious artifacts and ghosts around the edge of this shell.  

%\begin{figure*}
%	\centering
%	\includegraphics[width=1\linewidth,trim={7cm 1cm 7cm 2cm},clip]{MFFW_metric}
%	\caption{}
%	\label{fig:mffwmetric}
%\end{figure*}
%\begin{figure}
%	\centering
%	\includegraphics[width=1\linewidth,trim={4cm 0cm 4cm 1cm},]{MFFW3_metric}
%	\caption{}
%	\label{fig:mffw3metric}
%\end{figure}

%\begin{figure*}
%	\centering
%	\subfigure[BF]       {\includegraphics[width=0.13\linewidth]{rose_BF}}
%	\subfigure[BRW]      {\includegraphics[width=0.13\linewidth]{rose_BRW}}
%	\subfigure[CBF]      {\includegraphics[width=0.13\linewidth]{rose_CBF}}
%	\subfigure[CNN]      {\includegraphics[width=0.13\linewidth]{rose_CNN}}
%	\subfigure[CSR]      {\includegraphics[width=0.13\linewidth]{rose_CSR}}
%	\subfigure[DSIFT]    {\includegraphics[width=0.13\linewidth]{rose_DSIFT}}
%	\subfigure[GFDF]     {\includegraphics[width=0.13\linewidth]{rose_GFDF}}
%	\subfigure[GFF]      {\includegraphics[width=0.13\linewidth]{rose_GFF}}
%	\subfigure[IFCNN]    {\includegraphics[width=0.13\linewidth]{rose_IFCNN}}
%	\subfigure[MMFNet]   {\includegraphics[width=0.13\linewidth]{rose_MFFNet}}
%	\subfigure[MSTSR]    {\includegraphics[width=0.13\linewidth]{rose_MSTSR}}
%	\subfigure[MWGF]     {\includegraphics[width=0.13\linewidth]{rose_MWGF}}
%	\subfigure[SESF]     {\includegraphics[width=0.13\linewidth]{rose_SESF}}
%	\subfigure[Reference]{\includegraphics[width=0.13\linewidth]{rose_Reference}}
%	\caption{The fused images of No. 11 pair in MFFW2 dataset.}
%	\label{fig:rose}
%\end{figure*}

\subsection{Analysis}
Based on the experimental results, several interesting conclusions can be drawn: 

(i) Compared with classic methods, the improvement of current DL based models may be limited. From Tables \ref{tab:MFFW} and \ref{tab:MFFW3} and Figs. \ref{fig:coffee}-\ref{fig:shell}, it is observed that DL based models often perform worse than classic methods. CNN casts MFF task into a classification problem, where focus maps are regarded as the output. MMFNet employs a two-stage network, where the first and the second stages are used to generate initial focus maps and to refine them, respectively. Both IFCNN and SESF design auto-encoder based networks. Fusion strategies are implemented on codes (or say, feature maps). The reason why DL based methods perform badly on our datasets is that deep networks easily overfit on training set. However, as far as we known, all DL based methods exploit simulated training sets which are far away from real-world images. 

(ii) From our experiments, it is found that no method can achieve satisfactory results with regard to all testing pairs. In other words, their generalization capabilities are limited. The reasons are twofold: Firstly, most methods take a specific prior knowledge into their models. Nonetheless, owing to the complexity of real-world images, the specific prior knowledge cannot be applied to various cases. Secondly, except MMFNet, no method takes the DSE into account. It is not surprising that they perform badly, especially in the areas around boundaries of focus maps. In spite of MMFNet considering the DSE, it suffers from overfitting and breaks down in some cases (e.g., Fig. \ref{fig:coffee}). 

\section{Conclusion}
We construct a new multi-focus image fusion dataset called MFFW. The purpose of MFFW is to evaluate the performance of MFF algorithms in presence of DSE or complicated scenes. Our experiments show that SOTA methods on MFFW dataset cannot robustly generate satisfactory results. MFFW can be a new benchmark dataset for multi-focus image fusion, and MFFW will potentially boost the development of MFF field and reduce the gap between practice and MFF algorithms.

% or


%\appendices
%% you can choose not to have a title for an appendix
%% if you want by leaving the argument blank
%\section{}

%\section*{Acknowledgment}


\ifCLASSOPTIONcaptionsoff
  \newpage
\fi


\bibliographystyle{IEEEtran}
\bibliography{ref}

%%%%%%%%%%%%%%%%%%%%%%%%%%%%%%%%%%%%%%%%%%%%%%%%%%%%%%%%%%%%%%%%%%%%%%%%%%%%%
\vspace*{-5ex}%
\begin{biography}[{\includegraphics[width=1in,height=1.25in,clip,keepaspectratio]{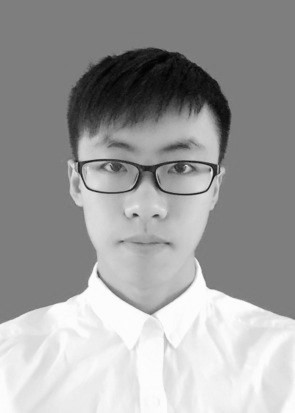}}]{Shuang Xu} is currently pursing the Ph.D. degree in statistics with the School of Mathematics and Statistics, Xi'an Jiaotong University, Xi'an, China. His current research interests include Bayesian statistics, deep learning and complex network.
\end{biography}
\vspace*{-5ex}%
%%%%%%%%%%%%%%%%%%%%%%%%%%%%%%%%%%%%%%%%%%%%%%%%%%%%%%%%%%%%%%%%%%%%%%%%%%%%%
\begin{biography}[{\includegraphics[width=1in,height=1.25in,clip,keepaspectratio]{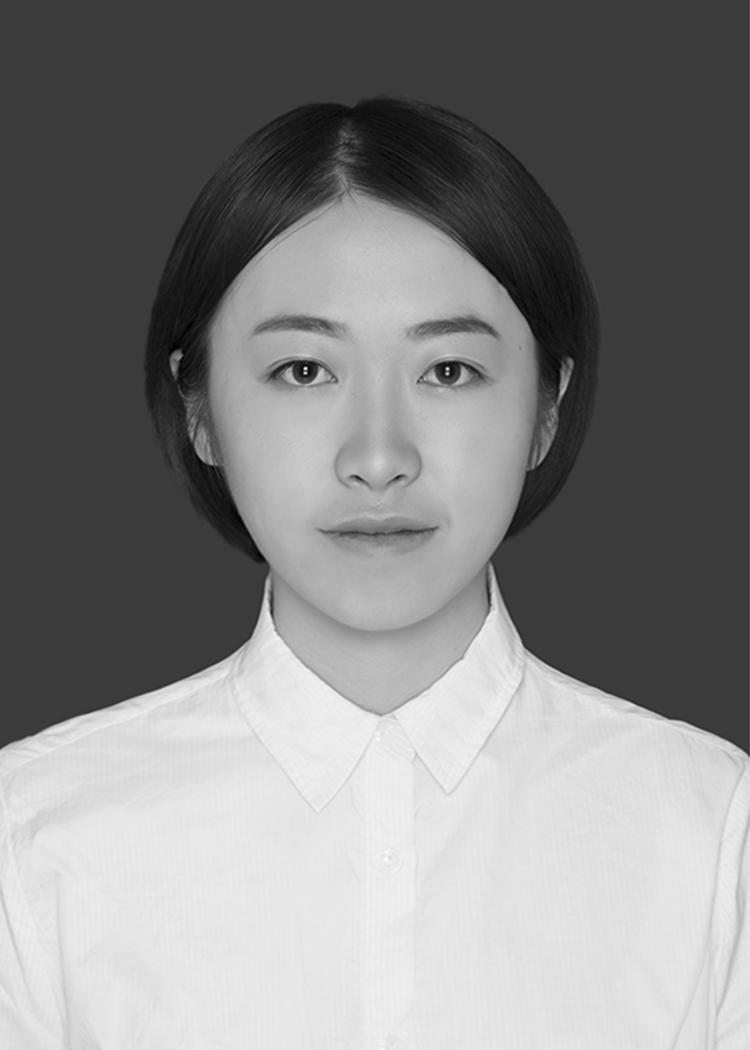}}]{Xiaoli Wei} 
	is currently pursing the master's degree in statistics with the school of Mathematics and Statistics, Xi'an Jiaotong University, Xi'an China. Her current research interests include classification, deep learning and seismic data fault detection.
\end{biography}
\vspace*{-5ex}%
\begin{biography}[{\includegraphics[width=1in,height=1.25in,clip,keepaspectratio]{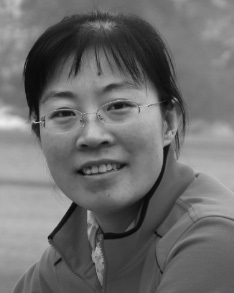}}]{Chun-Xia Zhang} received her Ph.D degree in Applied Mathematics from Xi'an Jiaotong University, Xi'an, 	China, in 2010.
	
	Currently, she is an associate professor in School of Mathematics and Statistics at Xi'an Jiaotong University. She has authored and coauthored about 30 journal papers on ensemble learning techniques, nonparametric regression and etc. Her main interests are in the area of ensemble learning, variable selection and deep learning.
\end{biography}
%%%%%%%%%%%%%%%%%%%%%%%%%%%%%%%%%%%%%%%%%%%%%%%%%%%%%%%%%%%%%%%%%%%%%%%%%%%%%
%%%%%%%%%%%%%%%%%%%%%%%%%%%%%%%%%%%%%%%%%%%%%%%%%%%%%%%%%%%%%%%%%%%%%%%%%%%%%
\vspace*{-5ex}%

\begin{biography}[{\includegraphics[width=1in,height=1.25in,clip,keepaspectratio]{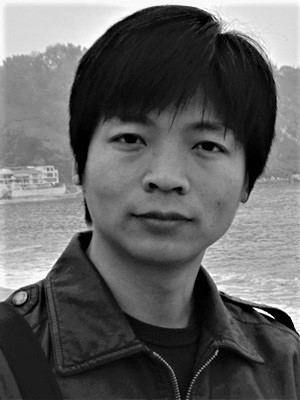}}]{Junmin Liu} (M'13)
	received the M.S. degree in computational mathematics from Ningxia University, Yinchuan, China, in 2009, and the Ph.D. degree in applied mathematics from Xi’an Jiaotong University, Xi’an, China, in 2013. 
	
	He is currently an Associate Professor with the School of Mathematics and Statistics, Xi’an Jiaotong University. His current research interests include hyperspectral unmixing, remotely sensed image
	fusion, and deep learning.
\end{biography}
%%%%%%%%%%%%%%%%%%%%%%%%%%%%%%%%%%%%%%%%%%%%%%%%%%%%%%%%%%%%%%%%%%%%%%%%%%%%%

%%%%%%%%%%%%%%%%%%%%%%%%%%%%%%%%%%%%%%%%%%%%%%%%%%
\vspace*{-5ex}%
\begin{biography}[{\includegraphics[width=1in,height=1.25in,clip,keepaspectratio]{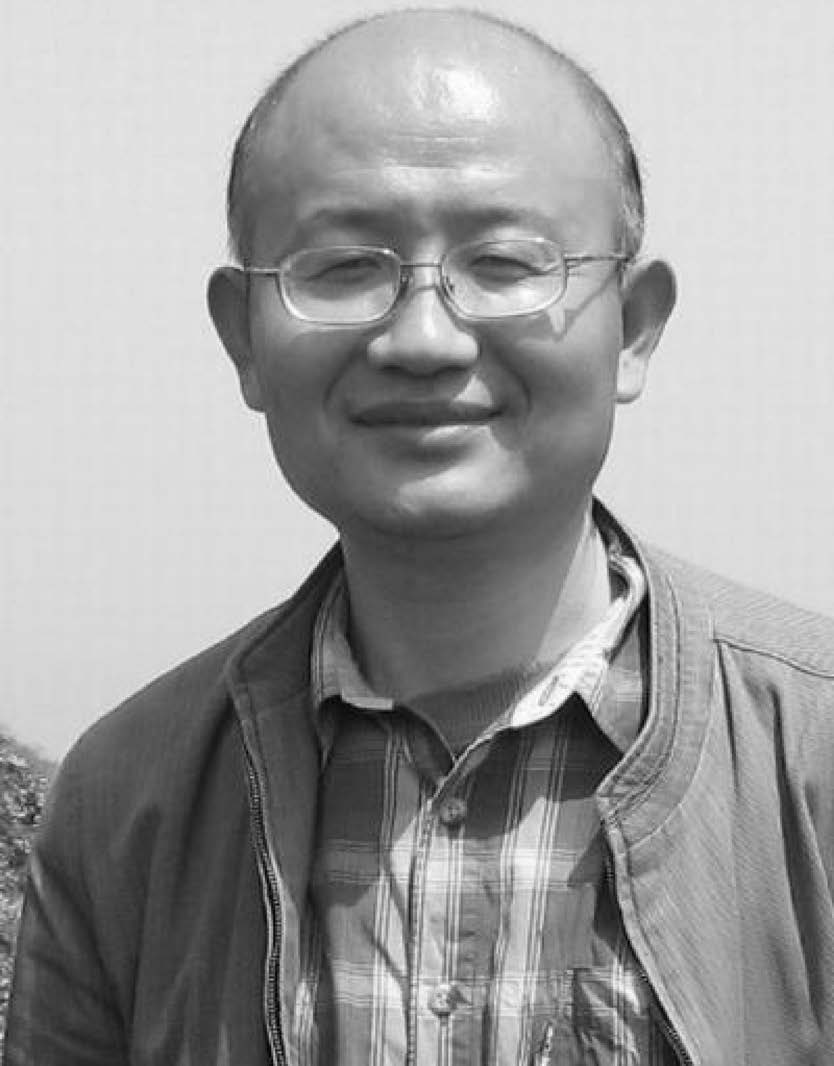}}]{Jiangshe Zhang} was born in 1962. He received the M.S. and Ph.D. degrees in applied mathematics from Xi'an Jiaotong University, Xi'an, China, in 1987 and 1993, respectively, where he is currently a Professor with the Department of Statistics. He has authored and co-authored one monograph and over 80 conference and journal publications on robust clustering, optimization, short-term load forecasting for electric power system, and remote sensing image processing. His current research interests include Bayesian statistics, global optimization, ensemble learning, and deep learning.
\end{biography}
%%%%%%%%%%%%%%%%%%%%%%%%%%%%%%%%%%%%%%%%%%%%%%%%%%%%%%%%%%%%%%%%%%

\end{document}